\documentclass{article}
\usepackage{iclr2027_conference,times}
\usepackage[T1]{fontenc}
\usepackage{amsmath,amssymb}
\usepackage{graphicx}
\usepackage{xcolor}
\usepackage{booktabs}
\usepackage{wrapfig}
\usepackage{multirow}
\usepackage{xspace}
\usepackage{colortbl}
\usepackage{caption}
\usepackage{enumitem}
\definecolor{cvprblue}{rgb}{0.21,0.49,0.74}
\usepackage[
  colorlinks=true,
  citecolor=cvprblue,
  urlcolor=cvprblue
]{hyperref}

\newcommand{\method}{TimelineControl\xspace}
\newcommand{\dataset}{TimelineMotion\xspace}

\usepackage{pifont}
\newcommand{\cmark}{\ding{51}} %
\newcommand{\xmark}{\ding{55}} %

\usepackage{url}
\usepackage{comment}
\usepackage{soul}
\hypersetup{hypertexnames=false}

\title{Streaming Multi-Track Timeline Control for 3D Human Motion Generation}

\author{%
\textbf{Yangsong Zhang$^{1,\dagger}$, Anujith Muraleedharan$^{1}$, Rikhat Akizhanov$^{1}$, G\"ul Varol$^{2}$,}\\
\textbf{Fabio Pizzati$^{1}$, Ivan Laptev$^{1}$} \\
\normalfont
$^{1}$Mohamed bin Zayed University of Artificial Intelligence (MBZUAI) \\
$^{2}$LIGM, \'Ecole des Ponts, IP Paris, Univ Gustave Eiffel, CNRS \\
$^{\dagger}$\texttt{yangsong.zhang@mbzuai.ac.ae}
}
\iclrpreprint

\begin{document}
\maketitle
\begin{figure}[!htbp]
    \vspace{-.5cm}
    \centering
    \includegraphics[width=\textwidth]{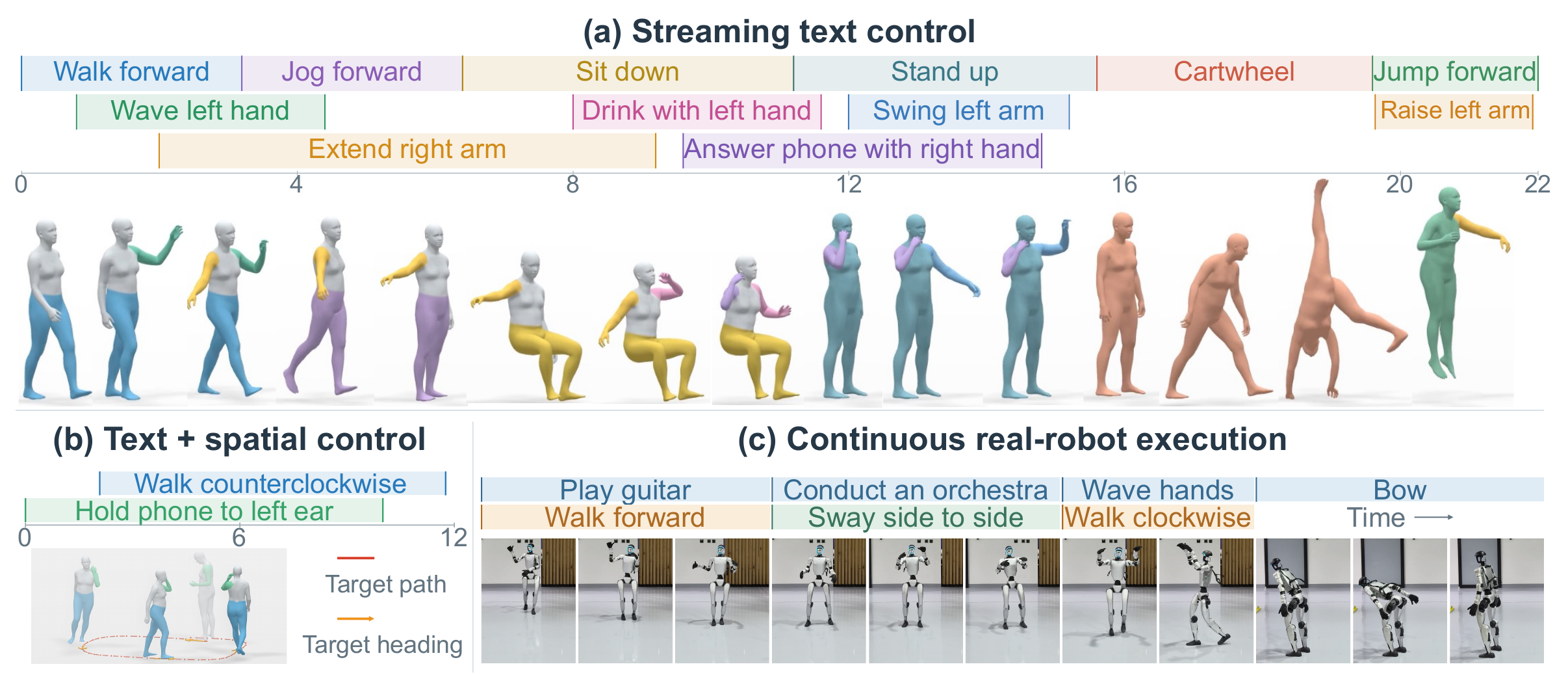}
    \caption{\textbf{Streaming multi-track motion generation with \method.} (a) Sequential and overlapping body-part instructions. (b) Joint text, root-path, and heading control. (c) Humanoid robot execution. 
    In (a) and (b), body-part colors correspond to instructions.
    }
    \label{fig:teaser}
\end{figure}

\begin{abstract}
Text-driven human motion generation has advanced substantially, yet most methods assume instructions are available before synthesis. Interactive applications require responding to new instructions while continuing ongoing actions, such as answering a phone while walking. Existing approaches address streaming generation or simultaneous composition without explicitly combining streaming instruction arrival with independently timed, overlapping actions. We introduce streaming multi-track timeline control and propose TimelineControl to incorporate new instructions alongside ongoing actions. Interval-aware conditioning preserves instruction timing, while causal part-structured representations and part-aware denoising coordinate concurrent actions across body regions. We also construct TimelineMotion, a dataset with overlapping instruction intervals and body-part annotations. Experiments on TimelineMotion and MTT demonstrate improved semantic alignment and temporal adherence over evaluated streaming baselines, including models retrained on the same data. Ablations and human evaluations validate our design, complemented by spatial conditioning and humanoid execution demonstrations. 
{\em Our code, data and models will become publicly available at \url{https://mael-zys.github.io/TimelineControl/}.}
\end{abstract}

\section{Introduction}
\label{sec:intro}

Text-driven human motion generation has made substantial progress in synthesizing realistic movements from natural language~\citep{petrovich2022temos,tevet2023mdm,chen2023mld,guo2024momask}, including action sequences and simultaneous motions~\citep{athanasiou2022teach,petrovich2024stmc,li2026frankenmotion}. However, most approaches assume that instructions are available before generation begins. Interactive applications in gaming and robotics require both reactivity and compositionality: a character may need to answer a ringing phone while walking, or avoid a moving obstacle while continuing another activity. Such interactions require incorporating new actions while preserving compatible ongoing behaviors and their individual timings.

This setting poses three coupled challenges: generating motion without access to future instructions, maintaining distinct temporal scopes for concurrent actions, and coordinating their effects across body regions while preserving whole-body coherence. Previous work addresses these challenges separately. Streaming methods such as MotionStreamer~\citep{xiao2025motionstreamer}, ActionPlan~\citep{nazarenus2026actionplan}, and ARDY~\citep{zhao2026ardy} support changing text conditions, but their described interfaces do not explicitly maintain independently timed, overlapping instructions. Conversely, SINC~\citep{athanasiou2023sinc}, STMC~\citep{petrovich2024stmc}, and FrankenMotion~\citep{li2026frankenmotion}  support simultaneous or temporal composition assume pre-specified conditions. %

We introduce streaming multi-track timeline control, where instructions arrive at their start times, retain separate temporal intervals, and may overlap with ongoing actions, as illustrated in Figure~\ref{fig:teaser}. Our framework, TimelineControl, extends ActionPlan with online multi-prompt conditioning and part-aware denoising. Instructions are encoded separately, with interval encoding and visibility masks determining their temporal applicability. A causal part-structured autoencoder provides anatomical motion latents, while text-derived relevance estimates guide each body region toward applicable instructions. We combine body-part and global features to produce coordinated whole-body motion. New instructions guide future motion without modifying previously generated motions.

For training and evaluation, we construct TimelineMotion with overlapping instruction intervals and body-part annotations through STMC-based composition of HumanML3D motions, followed by refinement and filtering. Experiments on TimelineMotion and MTT demonstrate strong semantic alignment and temporal adherence, outperforming evaluated streaming baselines, including ActionPlan retrained on the same data. Ablations validate our conditioning and part-aware components, while human evaluations favor our text consistency over ActionPlan and offline STMC. We further validate our method by showing strong single-prompt generation, spatial conditioning, and humanoid execution through a motion-tracking controller. Our main contributions are: \vspace{-.5cm}\\
\begin{itemize}[nosep,topsep=0pt,itemsep=2pt,leftmargin=1em]
    \item We introduce streaming multi-track timeline control and develop \textbf{\method} to incorporate newly arriving instructions alongside ongoing actions, combining interval-aware multi-prompt conditioning with part-aware motion generation.
    \item We construct \textbf{\dataset}, a dataset of motion sequences paired with overlapping instruction intervals and body-part annotations, supporting training and evaluation of streaming compositional generation.
    \item We demonstrate strong semantic alignment and temporal adherence on \dataset and MTT, supported by baseline comparisons, ablations, and human evaluations. We further demonstrate spatial conditioning and humanoid motion execution.
\end{itemize}

\section{Related Work}
\label{sec:related}

\noindent\textbf{Text-to-motion generation.} Text-to-motion methods represent data with raw motion~\citep{tevet2023mdm}, continuous latents~\citep{petrovich2022temos,chen2023mld} or discrete tokens~\citep{guo2022tm2t,zhang2023t2mgpt,zhang2026t2mgptplus}.
Generation strategies include autoregression~\citep{guo2022tm2t,zhang2023t2mgpt}, diffusion~\citep{tevet2023mdm,zhang2024motiondiffuse,chen2023mld}, masked token prediction~\citep{guo2024momask}, and masked autoregression~\citep{meng2025mardm,he2026molingo}.
For sequential composition, TEACH~\citep{athanasiou2022teach} conditions successive actions on motion history, while DoubleTake~\citep{shafir2024human} and FlowMDM~\citep{barquero2024flowmdm} synthesize transitions through coordinated denoising.
In discrete motion-token space, M2D2M~\citep{chi2024m2d2m} combines independent and joint denoising to compose action sequences.
For simultaneous actions, SINC~\citep{athanasiou2023sinc} constructs spatial compositions, while MultiAct~\citep{sala2026multiact} uses attention guidance to strengthen underrepresented actions in composite prompts.
MotionMaster~\citep{jiang2026motionmaster} combines quantized motion tokens with a pretrained multimodal language model for compositional generation and body-part editing.
STMC~\citep{petrovich2024stmc} brings temporal and spatial composition together by assigning text intervals to body regions and combining denoising predictions across tracks, given a complete timeline before synthesis.
Temporally localized text conditioning also appears in video generation~\citep{wu2025mint,shu2026tie}. None of these methods support streaming multi-track timeline control.

\noindent\textbf{Part-aware motion modeling.} Part-aware representations use hierarchical upper- and lower-body streams for action composition~\citep{ghosh2021sca} or body-part attention in the motion encoder~\citep{zhong2023attt2m}.
PB-MLD~\citep{liu2026pbmld} uses separate upper- and lower-body variational autoencoders with a joint Transformer denoiser.
SALAD~\citep{hong2025salad} preserves skeletal structure in its latent representation and combines temporal, skeletal, and text attention for generation and editing.
Part-specific generation uses inter-part communication in ParCo~\citep{zou2024parco}, part-specific text with full-body refinement in LGTM~\citep{sun2024lgtm}, and part-aware text grounding with holistic-part fusion in ParTY~\citep{heo2026party}.
Other approaches condition motion on spatial and temporal descriptions~\citep{zhang2023finemogen} or compose atomic motions for open-vocabulary generation~\citep{fan2026motiondoor}.
FineMotion~\citep{wu2025finemotion} provides detailed motion annotations, while FrankenMotion~\citep{li2026frankenmotion} pairs hierarchical temporal annotations with a diffusion model conditioned on pre-specified, temporally structured part descriptions.
We use six anatomical latents in a causal autoencoder to support part-wise denoising during streaming generation.\looseness=-1

\noindent\textbf{Streaming and interactive motion generation.} Causal tokenizers encode motion as discrete tokens~\citep{jiang2025motionstream} or continuous latents, as in MotionStreamer~\citep{xiao2025motionstreamer}.
Autoregressive diffusion models generate motion incrementally from motion history~\citep{shi2024amdm,chen2024camdm}.
Diffusion Forcing~\citep{chen2024diffusionforcing} introduces token-specific noise levels, which CMDM~\citep{yu2026cmdm} applies to causal motion generation and FloodDiffusion~\citep{cai2026flooddiffusion} adapts to changing text.
ActionPlan~\citep{nazarenus2026actionplan} jointly generates motion and frame-level action plans with latent-specific noise levels, while PRISM~\citep{ling2026prism} combines per-joint causal latents with flow matching and Self Forcing~\citep{huang2025selfforcing}.
We extend ActionPlan's joint formulation to prompts that arrive independently during generation and may overlap in time.
Spatial conditioning complements language through trajectories and keyframes~\citep{karunratanakul2023gmd}, joint-position constraints~\citep{xie2024omnicontrol}, and constraints for masked motion synthesis~\citep{pinyoanuntapong2025maskcontrol}.
DartControl~\citep{zhao2025dartcontrol} supports interactive motion control, while ARDY~\citep{zhao2026ardy} combines explicit root trajectories with body latents for online text and spatial conditioning.
Related settings include online two-character interaction in Ready-to-React~\citep{cen2025ready} and motion generation with physics-based character control in CLoSD~\citep{tevet2025closd}.
In this paper, we mainly focus on online streaming timeline control and treat spatial targets as optional conditions on the timeline generator.

\section{Preliminaries}
\label{sec:preliminaries}

\paragraph{Prompt-conditioned rectified flow.}
We follow~\citet{nazarenus2026actionplan} for our problem setup.
Given a text prompt $\ell$, we generate a latent motion sequence
$z=(z_1,\ldots,z_N)$, where $z_n$ represents the $n$-th motion segment, $n\in[1, N]$. The generator is synthesizing $N$ latent motions, then decoded to poses. The latent space is defined by a pretrained motion autoencoder that remains
frozen during generator training. The generated latent sequence is later
decoded back to motion~\citep{chen2023mld,xiao2025motionstreamer,nazarenus2026actionplan}. Let $z^0$ denote a clean latent sequence and $t\in[0,1]$ the
rectified-flow time~\citep{liu2023rectifiedflow}. We construct the noisy latent $z^t$ and target
velocity $v^\star$ as
\begin{equation}
    z^t = (1-t)z^0 + t\epsilon,
    \qquad
    v^\star = \epsilon - z^0,
    \qquad
    \epsilon \sim \mathcal{N}(0,\mathbf{I}).
\label{eq:rectified-flow}
\end{equation}
Here, $t=0$ corresponds to clean motion and $t=1$ to Gaussian noise. Then, we use flow matching~\citep{lipman2023flowmatching} to train a velocity predictor $v_\theta$, with parameters $\theta$,
conditioned on $\ell$:
\begin{equation}
    \mathcal{L}_{\mathrm{flow}}
    =
    \mathbb{E}_{(z^0,\ell),\,t,\,\epsilon}
    \left[
        \left\|
        v_\theta(z^t,t,\ell)-v^\star
        \right\|_2^2
    \right].
\label{eq:flow-objective}
\end{equation}

At inference, we initialize $z^1\sim\mathcal{N}(0,\mathbf{I})$ and
progressively transport it toward $t=0$. Let $\widehat{z}^t$ be the
current sample and $t'<t$ the next flow time. We obtain the next latent $\hat{z}^{t'}$ with a first-order update:
\begin{equation}
    \widehat{z}^{t'}
    \leftarrow
    \widehat{z}^{t}
    +
    (t'-t)\,
    v_\theta(\widehat{z}^t,t,\ell),
    \qquad
    0\leq t'<t\leq 1.
\label{eq:flow-inference}
\end{equation}
When $t=0$, the resulting latent sequence $\widehat{z}^0$ is decoded
into motion.

\paragraph{Online multi-track timeline control.}
We follow ActionPlan~\citep{nazarenus2026actionplan} for the streaming setup, but we also assume that motion instructions
\textit{may arrive at arbitrary times, including when another
instruction is not completed}. We aim to generate motion of each instruction in the correct interval. An instruction $i$ is represented as
$e_i=(\ell_i,[s_i,u_i))$, where $\ell_i$ is a textual prompt describing
the action and $s_i,u_i$ are start and end interval positions in the same window-local latent-index coordinates as $n$. During streaming,
$z$ belongs to the \textit{current denoising window}, containing
previously emitted motion as fixed history and a continuation to be
generated. Thus, $N$ counts latent entries within this window, with index $n$. At a streaming step $\tau$, the received instructions overlapping the current denoising window $\mathcal W_\tau$ form $C_\tau = \{e_i \mid s_i \le \tau,\; [s_i,u_i) \cap \mathcal{W}_\tau \neq \varnothing\}$, where $\mathcal{W}_\tau$ is the current denoising window. The streaming time $\tau$ is expressed in the same window-local latent-index coordinates as $n$, $s_i$, and $u_i$; these coordinates are updated whenever the window advances. Let $n_\tau$ denote the
first of the $N$ latent entries to be generated in causal generation:
$z_{<n_\tau}$ is retained as context, while
$z_{n_\tau:N}$ is to be generated. The denoiser
$v_\theta(z^t,t,\mathcal{C}_\tau)$ processes the whole window, but
noise updates apply only to the continuation. As completed motion is emitted,
it becomes fixed history and the window advances. When a new
instruction arrives at time $\tau'$, we update the conditions to
$\mathcal{C}_{\tau'}$ and regenerate the continuation $z_{n_{\tau'}:N}$ in the current window. We now describe how we enable the estimation of $v_\theta(z^t,t,\mathcal{C}_\tau)$.\looseness=-1

\section{Method}
\label{sec:method}

\begin{figure}[t]
    \centering
    \includegraphics[width=\textwidth]{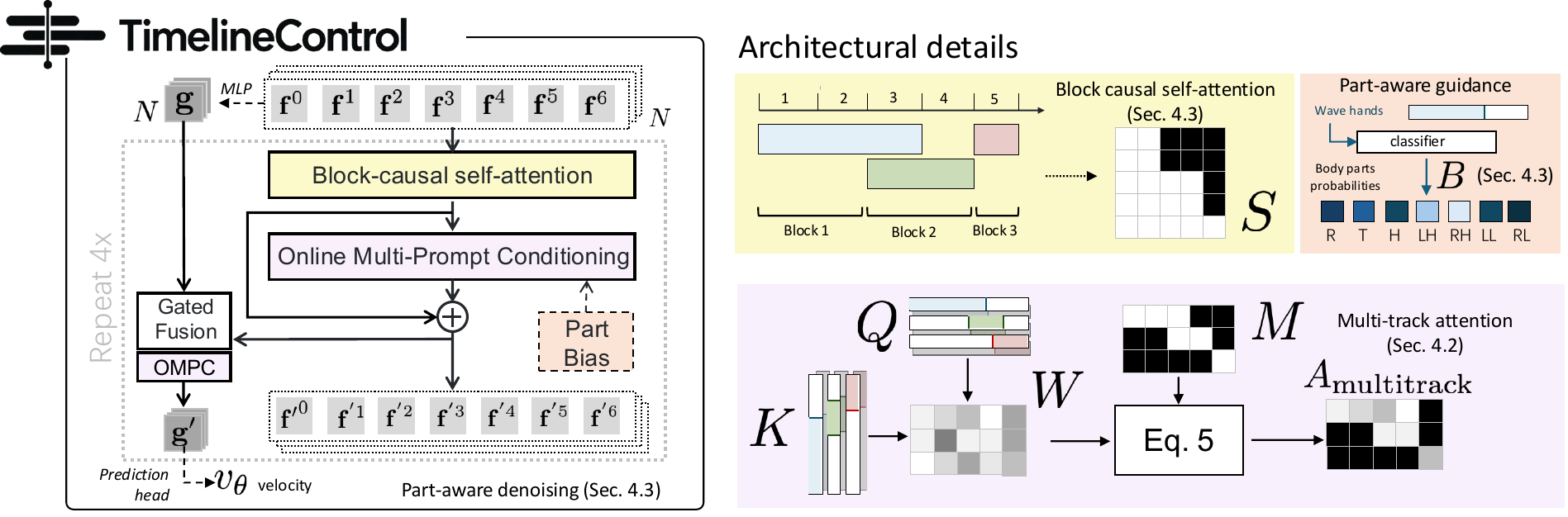}
    \caption{\textbf{\method.} We process part features and whole-body features separately (left). The part features undergo block-causal self-attention (yellow). Both use online multi-prompt conditioning (pink, OMPC) to allow for multi-track streaming inference. We add a part bias (orange) to enforce better mapping to body parts. Details (right) show examples of elements used in text.\looseness=-1}
    \label{fig:model_overview}
\end{figure}

We introduce \method and show its architecture in Fig.~\ref{fig:model_overview}. Our method is built on ActionPlan~\citep{nazarenus2026actionplan}, and trained with Diffusion Forcing~\citep{chen2024diffusionforcing} to impose streaming generation. To allow multi-track conditioning, we use online multi-prompt conditioning, aligning
motion features with available streamed prompts (Sec.~\ref{sec:online-conditioning}). We also find that
part-aware representations help disambiguate overlapping instructions, as different instructions may impact different body parts. We exploit this with a part-structured autoencoder (Sec.~\ref{sec:autoencoder}) and
with part-aware and global features (Sec.~\ref{sec:part-aware-denoising}) to guide individual regions while maintaining whole-body awareness.

\subsection{Causal Part-Structured Autoencoder}
\label{sec:autoencoder}

\paragraph{Part-structured representation.}
For multi-track control, we assume that different instructions may impact different parts of the body. Hence, we benefit from part-structured representations in the latent space in which the motion generator is operating, similarly to related works~\citep{heo2026party,hong2025salad}.
In practice, we partition body motion into six regions: torso, head, left and right arms, and left and right legs. Each entry $z_n$ of the sequence
introduced in Section~\ref{sec:preliminaries} contains six body latents $b_n^k \in \mathbb{R}^8$, where
$k \in {1,\ldots,6}$ indexes these regions. Following
ARDY~\citep{zhao2026ardy}, root motion bypasses the body encoder, keeping global position and heading explicit for spatial conditioning.
The three position coordinates and the sine and cosine of heading
over the same four frames form $r_n \in \mathbb{R}^{20}$.
The motion representation is $z_n=[r_n;b_n^1;\ldots;b_n^6]\in\mathbb{R}^{68}$,
with $z\in\mathbb{R}^{N\times68}$.\looseness=-1

\paragraph{Architecture.}
We define our autoencoder for causal streaming-oriented generation. The encoder and decoder each comprise three causal temporal-convolution
stages, compressing the motion by a factor of 4. Within these stages, skeleton-graph message passing connects the torso stream to the head and each limb~\citep{hong2025salad}. Temporal convolutions capture each region's motion history, while graph communication provides context from connected regions without merging their representations. The decoder additionally conditions on root yaw rate, horizontal velocity, and height derived from the explicit trajectory. Following MotionStreamer~\citep{xiao2025motionstreamer}, we train
the autoencoder with reconstruction and KL losses, and add additional acceleration loss to reduce jitter, then freeze it during generator training. The detailed architecture is illustrated in Appendix~\ref{app:autoencoder}.

\subsection{Online Multi-Prompt Conditioning}
\label{sec:online-conditioning}

\paragraph{Interval-aware text attention.}
To implement the conditioning $v_\theta(z^t,t,\mathcal{C}_\tau)$,
we must allow different instructions to influence different portions
of the motion, with overlaps. We assume text cross attention for conditioning. Let $F=[\mathbf{f}_1;\ldots;\mathbf{f}_N]$ be the hidden features
entering the cross-attention layer. These are
computed from the noisy motion input $z^t$ by the preceding layers. In the first layer, they are $z_n$. Each row $\mathbf{f}_n$ retains the
temporal index $n$ of the motion sequence and supplies the query for
that position. Section~\ref{sec:part-aware-denoising} specifies how $F$ evolves inside the network.

We encode each prompt $\ell_i\in\mathcal{C}_\tau$ independently with a frozen
CLIP~\citep{radford2021clip} text encoder and stack the resulting sentence embeddings $h_i$
into $H=[h_1;\ldots;h_J]$, where $J=|\mathcal{C}_\tau|$ is the number
of received instructions for a streaming window. For one attention head of dimension $d$,
learned projections map $F$ to queries $Q\in\mathbb{R}^{N\times d}$
and $H$ to keys and values \textit{of all the instructions, for all $N$ steps}: $K,V\in\mathbb{R}^{J\times d}$.
To align motion features with each instruction's temporal scope,
we apply rotary position encoding (RoPE)~\citep{su2024roformer} to queries using their
motion positions and rotary interval encoding (RoTE) to keys using
each instruction's midpoint and duration, as in TIE~\citep{shu2026tie}.
We then define the score matrix
$W=\operatorname{RoPE}(Q)\operatorname{RoTE}(K)^\top/\sqrt{d}
\in\mathbb{R}^{N\times J}$, which is the standard dot-product
attention without softmax normalization. Each of its $J$ columns
corresponds to a distinct prompt, so every motion position is
scored separately and aggregated in $W$. In short, we extract relevance scores \textit{for each prompt in the input batch} with the current motion features.\looseness=-1

\vspace{-3px}\paragraph{Multi-track attention.}
Receiving
an instruction does not make it applicable to the entire denoising
window: it should contribute only between its start and end times.
Positional encoding supplies timing bias but does not enforce
this restriction. We hence define a visibility matrix
$M\in\{0,1\}^{N\times J}$:\vspace{-2px}
\begin{equation}
    M_{ni} =
    \begin{cases}
        1, & e_i\in\mathcal{C}_\tau
             \ \text{and}\ s_i\leq n<u_i,\\
        0, & \text{otherwise}.
    \end{cases}
\label{eq:prompt_mask}
\end{equation}
We show $M$ in Fig.~\ref{fig:model_overview}. At streaming time $\tau$, this permits direct text contributions only
from instructions whose intervals cover the queried motion.
Hence, different rows can admit different subsets of the same
instructions. Let $c_n=\sum_i M_{ni}$ count the visible instructions
for the latent position $n$, and let the column vector $\mathbf{c}$ collect these
counts. Whereas $W$ contains pairwise compatibility scores, the final
attention matrix $A_{\mathrm{multitrack}} \in \mathbb{R}^{N \times J}$ determines how much each
instruction contributes to each group of generated frames. Given an
optional bias matrix $B\in\mathbb{R}^{N\times J}$, defined in
Sec.~\ref{sec:part-aware-denoising}, we have:\looseness=-1\vspace{-4px}
\begin{equation}
    A_{\mathrm{multitrack}}
    =
    \frac{
        M\odot\sigma(W+B)
    }{
        \sqrt{\max(1,\mathbf{c})}
    }.
\label{eq:multitrack-attention}
\end{equation}\vspace{-4px}

\vspace{-4px}Here, $\sigma$ is sigmoid and $\odot$ is element-wise multiplication. We use sigmoid attention~\citep{ramapuram2409theory} to preserve instruction importance. $B$ adjusts part relevance, $M$
excludes inapplicable instructions, and the normalization limits the text contribution in case of overlap. The resulting cross-attention is
$\operatorname{CA}_{\mathrm{multitrack}}(F,H;B,M)=A_{\mathrm{multitrack}}V \in \mathbb{R}^{N \times d}$. Note in this operation, aggregation over timelines happens.
The responses from multiple heads are concatenated and
projected back to $F$ dimensionality. Sec.~\ref{sec:part-aware-denoising} explains how we integrate this in \method.

\subsection{Part-Aware Denoising}
\label{sec:part-aware-denoising}
\paragraph{Parts and whole body.}
We now describe the flow of features $F$ inside \method, and illustrate that in Fig.~\ref{fig:model_overview} (left). Multi-track control requires different body parts to follow overlapping
instructions without losing whole-body coordination. For it, we join part-aware feature processing with a global stream mixing information across different parts. At the first layer, a multilayer perceptron initializes
the whole-body features as
$\mathbf{g}_n=\operatorname{MLP}([\,\mathbf{f}_n^0;\ldots;\mathbf{f}_n^6\,])$. This combines information
across the root and six body regions while preserving their separate
part-aware information. 

\vspace{-3px}\paragraph{Block-causal self-attention.}
Then, we process $F$ with self-attention. During training, when instruction $e_i$ arrives at streaming time $\tau=s_i$, it defines
a generation block
$\mathcal{B}_\tau=\{n\in\{1,\ldots,N\}:\tau\le n<\tau^+\}$,
where $\tau^+$ is the next distinct arrival position within the window,
or $N+1$ if there is none.
Motion within a block is denoised jointly using the instructions
available in $\mathcal{C}_\tau$, allowing bidirectional attention within
the block and attention to all previous motion.
Bidirectional attention coordinates motion within
the block, so that earlier poses can anticipate later movements and
form a temporally coherent action.
For each query position $n\in\mathcal{B}_\tau$ and key position $m$,
we add a mask to the self-attention logits, with
$S^{(\tau)}_{nm}=0$ if $m<\tau$ or $m\in\mathcal{B}_\tau$,
and $S^{(\tau)}_{nm}=-\infty$ otherwise.
We \textit{only enforce this at training time} to improve learning.
We show an example of $S$ in Fig.~\ref{fig:model_overview}.

\vspace{-3px}\paragraph{Part-aware guidance.}
After self-attention, the features then receive the residual contribution of
$\operatorname{CA}_{\mathrm{multitrack}}$ from
Section~\ref{sec:online-conditioning}, using the temporally updated
features as queries and $B^k$ as guidance. To guide each part toward relevant instructions, a text classifier is
trained jointly with the denoiser using dataset body-part labels obtained
through an LLM. See Appendix~\ref{app:data_construction} for details. For all $J$ received instructions, it
predicts probabilities $\Pi\in[0,1]^{J\times6}$, with one column per
body part. For part $k$, the attention bias is
$B^k_{ni}=\lambda_{\mathrm r}(\Pi_{ik}-\bar\Pi_i)$, where
$\bar\Pi_i$ averages instruction $i$'s six probabilities and
$\lambda_{\mathrm r}$ controls guidance strength. Repeating these biases
over the $N$ motion positions gives $B^k\in\mathbb{R}^{N\times J}$.
The root uses $B^0=0$. Classifier predictions are detached when biasing
attention, while temporal applicability remains governed by $M$. See Fig.~\ref{fig:model_overview} for a visualization.

\vspace{-3px}\paragraph{Gated part-global fusion} We finally fuse part-aware processed features with the global branch. Given the part-aware states
$\mathbf{f}_n^{\prime k}$, we fuse them into the
whole-body state: $\overline{\mathbf{g}}_n=\mathbf{g}_n+\alpha_n\,\operatorname{Attn}\!\left(
        \mathbf{g}_n,
        \{\mathbf{f}_n^{\prime k}\}_{k=0}^{6}
    \right),$
where the learned gate $\alpha_n\in[0,1]$ controls the contribution and $\operatorname{Attn}$ is cross-attention.
The whole-body state supplies the query, and the updated part-aware
states supply keys and values. The fused features
$\overline{\mathbf{g}}_n$ then pass through four Transformer layers
with block-causal self-attention and online text attention.
Whole-body text attention uses $B=0$ and replaces the normalization
of Section~\ref{sec:online-conditioning} with the row-wise sum of
masked sigmoid weights~\citep{ramapuram2409theory}.
After four stages, an output head maps the global features to the velocity.

\subsection{Training data}\label{sec:data}

Since the most relevant timeline dataset MTT~\citep{petrovich2024stmc} has only testing data and there is no training data for multi-track streaming motion generation, we construct \dataset, a dataset specific for the task of 44,944 motion sequences with overlapping text intervals, from HumanML3D~\citep{guo2022humanml3d} using STMC-based motion composition~\citep{petrovich2024stmc}.
Each timeline uses source clips from a single HumanML3D split. We first use Qwen3-8B~\citep{qwen3} to extract impacted body parts and motion attributes for HumanML3D captions. 
We then compose text labels in multiple tracks, using annotated labels for assessing motion compatibility. For instance, it is not feasible to simultaneously clap and point a finger with the left hand, because both actions impact that hand. Composition of different instructions can be sequential or parallel. The body part labels are also used to train the classifier in Section~\ref{sec:part-aware-denoising}. We then generate motion associated to the composed instructions. We first initialize non-overlapping intervals from their HumanML3D source-motion crops. Within overlapping intervals, instead, we use STMC to perform body-part stitching and temporal composition to combine source motions according to the annotated body-part assignments. Composing motions with STMC from different clips can potentially introduce abrupt changes around the boundaries. We refine these transition regions to improve continuity, by applying a similar refinement pipeline to InterPose~\citep{zhang2026interpose}. In short, we check for discontinuities by thresholding joints' position and rotation differences across frames, and smooth detected discontinuities with interpolation. Finally, we also discard sequences which still present abrupt translations or excessive jitter. Further dataset construction details are provided in Appendix~\ref{app:data_construction}.\looseness=-1

\section{Experiments}
\label{sec:experiments}

\subsection{Experimental Setup}
\label{subsec:setup}

\noindent\textbf{Datasets and baselines.}
We evaluate on the test sets of timeline dataset \dataset and MTT~\citep{petrovich2024stmc}, the most relevant dataset with multi-track instructions available. We further evaluate the performance on non-streaming text dataset HumanML3D~\citep{guo2022humanml3d}. 
Note \textit{no baseline natively supports streaming multi-track conditioning}. For streaming baselines, we compare with ActionPlan~\citep{nazarenus2026actionplan}, which also serves as our backbone, MotionStreamer~\citep{xiao2025motionstreamer}, DartControl~\citep{zhao2025dartcontrol}, and ARDY~\citep{zhao2026ardy}.
Our model and all baselines marked with $^{*}$ are trained jointly on \dataset and HumanML3D. Streaming methods receive each prompt at its annotated start time.
For baselines without native conditioning on multiple text intervals, we update a single caption at every instruction start and end: one active prompt is passed unchanged, and multiple active prompts are joined with \emph{``while''}.
Retrained baselines also use \emph{``while''} to join overlapping descriptions within each training window.
STMC~\citep{petrovich2024stmc} and FrankenMotion~\citep{li2026frankenmotion} are evaluated offline with the complete timeline with explicit body-part labels and body text intervals, respectively.
FrankenMotion additionally receives a sequence-level caption formed from the original event descriptions. On HumanML3D, it receives only the sequence-level caption.

\noindent\textbf{Evaluation metrics.}
We use TMR-based R-Precision, text-motion similarity (M2T), and similarity to source (M2M) to assess sequence and body-part alignment following~\citep{petrovich2023tmr,li2026frankenmotion}.
We use FID~\citep{heusel2017gans} to compare distributions of complete generated and reference motions, also including FID+ following STMC~\citep{petrovich2024stmc}.
Transition distance (Trans)~\citep{petrovich2024stmc} assesses continuity at instruction boundaries and it is reported in centimeters. Autoencoder reconstruction is evaluated with FID and mean per-joint position error (MPJPE, in millimeters) following ~\citet{xiao2025motionstreamer}.
We also measure if generated motions express each instruction within its requested interval. To do so, we score each instruction against sliding motion windows across the complete timeline using TMR embeddings. We then quantify with average precision (AP) and area-under-the-curve (AUC) if windows centered inside the requested instruction interval receive higher scores than those outside. More details in Appendix~\ref{app:evaluation_protocols}.

\noindent\textbf{Training/inference.}
We train jointly on \dataset and HumanML3D and each timeline window contains up to 78 prediction latents and 8 clean history latents.
On a NVIDIA RTX 6000 Ada at batch size one, sampling takes 280\,ms for the first motion token and 35\,ms per subsequent token. We also use the plan branch of ActionPlan unchanged from theirs. Full details are in App.~\ref{app:objectives}.

\subsection{Comparison with baselines}

\begin{table}[!t]
    \centering
    \caption{\textbf{Benchmarks with streaming instructions.} We outperform all the baselines on fine-grained evaluation about sequence alignment, body parts, motion quality and temporal adherence with the sequence, on both \dataset and MTT.
    }
    \label{tab:main}

    \small
    \setlength{\tabcolsep}{3pt}
    \renewcommand{\arraystretch}{1.05}

    \resizebox{\linewidth}{!}{%
    \begin{tabular}{@{}ll*{14}{c}@{}}
        \toprule

        & \multirow[c]{2}{*}{\textbf{Method}}
        & \multirow[c]{2}{*}{\textbf{Stream}}
        & \multicolumn{5}{c}{\textbf{Sequence alignment}}
        & \multicolumn{3}{c}{\textbf{Body-part alignment}}
        & \multicolumn{3}{c}{\textbf{Motion quality}}
        & \multicolumn{2}{c}{\textbf{Temporal adherence}}
        \\

        \cmidrule(lr){4-8}
        \cmidrule(lr){9-11}
        \cmidrule(lr){12-14}
        \cmidrule(lr){15-16}

        &
        &
        &
        R@1$_{32}$ $\uparrow$
        & R@2$_{32}$ $\uparrow$
        & R@3$_{32}$ $\uparrow$
        & M2T $\uparrow$
        & M2M $\uparrow$
        & R@3$_{32}$ $\uparrow$
        & M2T $\uparrow$
        & M2M $\uparrow$
        & FID $\downarrow$
        & FID+ $\downarrow$
        & Trans $\downarrow$
        & AP $\uparrow$
        & AUC $\uparrow$
        \\

        \midrule

        \multirow[c]{8}{*}{%
            \rotatebox[origin=c]{90}{\textbf{\textit{\dataset}}}}

        & STMC
        & \xmark
        & \underline{0.329}
        & \underline{0.452}
        & \underline{0.525}
        & \underline{0.627}
        & \underline{0.630}
        & \underline{0.353}
        & 0.538
        & 0.633
        & 0.055
        & 0.228
        & \underline{0.92}
        & \underline{0.803}
        & \underline{0.732}
        \\

        & FrankenMotion
        & \xmark
        & 0.106 & 0.178 & 0.232 & 0.538 & 0.551
        & 0.270 & 0.526 & 0.568
        & 0.159 & 0.246 & 1.01
        & 0.639 & 0.545
        \\

        & ActionPlan
        & \cmark
        & 0.248 & 0.344 & 0.405 & 0.585 & 0.585
        & 0.303 & \underline{0.547} & 0.589
        & 0.274 & 0.330 & 2.16
        & 0.733 & 0.653
        \\

        & ActionPlan$^{*}$
        & \cmark
        & 0.320 & 0.436 & 0.502 & 0.622 & 0.627
        & 0.347 & 0.546 & \underline{0.634}
        & 0.055 & \underline{0.209} & \underline{0.92}
        & 0.784 & 0.711
        \\

        & MotionStreamer$^{*}$
        & \cmark
        & 0.222 & 0.321 & 0.383 & 0.585 & 0.592
        & 0.304 & 0.541 & 0.608
        & \underline{0.049} & 0.218 & 1.09
        & 0.734 & 0.653
        \\

        & DartControl$^{*}$
        & \cmark
        & 0.263 & 0.381 & 0.454 & 0.604 & 0.575
        & 0.341 & 0.541 & 0.591
        & 0.244 & 0.368 & 1.27
        & 0.793 & 0.728
        \\

        & ARDY
        & \cmark
        & 0.165 & 0.246 & 0.303 & 0.562 & 0.575
        & 0.275 & 0.524 & 0.573
        & 0.340 & 0.383 & \textbf{0.90}
        & 0.697 & 0.607
        \\

        & \method
        & \cmark
        & \textbf{0.406}
        & \textbf{0.531}
        & \textbf{0.603}
        & \textbf{0.653}
        & \textbf{0.656}
        & \textbf{0.374}
        & \textbf{0.553}
        & \textbf{0.659}
        & \textbf{0.017}
        & \textbf{0.192}
        & 1.31
        & \textbf{0.832}
        & \textbf{0.772}
        \\

        \midrule

        \multirow[c]{8}{*}{%
            \rotatebox[origin=c]{90}{\textbf{\textit{MTT}}}}

        & STMC
        & \xmark
        & 0.330 & 0.491 & 0.566
        & \underline{0.678}
        & \underline{0.668}
        & 0.421 & 0.560 & \textbf{0.654}
        & -- & \textbf{0.476} & \textbf{0.49}
        & 0.769 & 0.729
        \\

        & FrankenMotion
        & \xmark
        & 0.140 & 0.234 & 0.309 & 0.572 & 0.577
        & 0.370 & 0.549 & 0.578
        & -- & 0.543 & 1.08
        & 0.629 & 0.579
        \\

        & ActionPlan
        & \cmark
        & 0.296 & 0.429 & 0.514 & 0.645 & 0.628
        & 0.399 & 0.551 & 0.615
        & -- & 0.589 & 1.01
        & 0.754 & 0.716
        \\

        & ActionPlan$^{*}$
        & \cmark
        & \underline{0.333}
        & \underline{0.497}
        & \underline{0.574}
        & \underline{0.678}
        & 0.656
        & 0.421
        & 0.560
        & \underline{0.633}
        & --
        & 0.535
        & 0.64
        & 0.778
        & 0.742
        \\

        & MotionStreamer$^{*}$
        & \cmark
        & 0.215 & 0.357 & 0.432 & 0.612 & 0.615
        & 0.358 & 0.543 & 0.612
        & -- & 0.489 & 0.90
        & 0.710 & 0.670
        \\

        & DartControl$^{*}$
        & \cmark
        & 0.311 & 0.449 & 0.543 & 0.651 & 0.614
        & \underline{0.426}
        & \underline{0.567}
        & 0.594
        & --
        & 0.604
        & 1.16
        & \underline{0.792}
        & \underline{0.764}
        \\

        & ARDY
        & \cmark
        & 0.207 & 0.321 & 0.369 & 0.598 & 0.601
        & 0.320 & 0.521 & 0.583
        & -- & 0.687 & \underline{0.57}
        & 0.674 & 0.621
        \\

        & \method
        & \cmark
        & \textbf{0.368}
        & \textbf{0.543}
        & \textbf{0.628}
        & \textbf{0.696}
        & \textbf{0.682}
        & \textbf{0.433}
        & \textbf{0.569}
        & \textbf{0.654}
        & --
        & \underline{0.478}
        & 0.68
        & \textbf{0.809}
        & \textbf{0.777}
        \\

        \bottomrule
    \multicolumn{16}{c}{\small{Methods with $*$ are trained on both HumanML3D and \dataset}}

    \end{tabular}%
    }
\end{table}

\noindent\textbf{Quantitative results.}
As presented in Table~\ref{tab:main}, \method achieves state-of-the-art performance on the two benchmarks supporting streaming inferences. Methods supporting online streaming inference are indicated in the Stream column.
Notably, \method also outperforms offline STMC on \dataset, despite STMC having access to the complete timeline and being used to compose our training data. For example, FID decreases from 0.055 to 0.017. 
This shows that training on STMC-composed data does not limit \method to the performance of the composition procedure.
Our data construction strategy alone helps with the task. Indeed, retraining ActionPlan jointly on \dataset and HumanML3D (ActionPlan$^*$ in Table~\ref{tab:main}) improves almost all the metrics on the two timeline benchmarks. 
However, our gains in sequence alignment and body-part alignment persist against retrained baselines, showing the impact of our contributions.
Appendix~\ref{app:actionplan_multitext} further compares with ActionPlan, our backbone, with separately encoded instructions and interval masks.\looseness=-1

We include HumanML3D in our evaluation to assess whether our model preserves performance in the standard single-prompt setting. As shown in Table~\ref{tab:main_humanml3d}, \method 
achieves the highest retrieval performance (0.874 vs 0.830) and the second-best FID among the evaluated baselines (0.027 vs 0.022).
These results suggest that its improvements on timeline benchmarks do not significantly impact text-to-motion generation.\begin{wraptable}{r}{0.5\linewidth}
\vspace{4px}
\centering
\caption{\textbf{Results on non-streaming generation.} We compare with motion generation methods on HumanML3D dataset. 
}
\label{tab:main_humanml3d}
\small
\setlength{\tabcolsep}{1.3pt}
\resizebox{\linewidth}{!}{%
\begin{tabular}{c@{\hspace{3pt}}lcccccc}\toprule
& \textbf{Method}
& R@1$_{32}$ $\uparrow$
& R@2$_{32}$ $\uparrow$
& R@3$_{32}$ $\uparrow$
& M2T $\uparrow$
& M2M $\uparrow$
& FID $\downarrow$ \\
\midrule
\multirow{8}{*}{\rotatebox{90}{\textbf{\textit{HumanML3D\hspace{10px}}}}}
& STMC & 0.611 & 0.756 & 0.815 & \underline{0.741} & 0.697 & 0.224 \\
& FrankenMotion & 0.432 & 0.566 & 0.643 & 0.677 & 0.689 & 0.079 \\
& ActionPlan & 0.618 & 0.768 & 0.829 & 0.724 & \textbf{0.769} & \textbf{0.022} \\
& ActionPlan$^{*}$ & \underline{0.645} & \underline{0.778} & \underline{0.830} & 0.734 & \underline{0.765} & 0.044 \\
& MotionStreamer$^{*}$ & 0.568 & 0.706 & 0.772 & 0.726 & 0.742 & 0.047 \\
& DartControl$^{*}$ & 0.536 & 0.699 & 0.775 & 0.711 & 0.657 & 0.345 \\
& ARDY & 0.413 & 0.566 & 0.649 & 0.657 & 0.649 & 0.349 \\
& \method & \textbf{0.687} & \textbf{0.827} & \textbf{0.874} & \textbf{0.769} & 0.753 & \underline{0.027} \\
\bottomrule
\end{tabular}%
}
\vspace{-15px}
\end{wraptable}

\begin{figure}[t]
    \centering
    \includegraphics[width=\linewidth]{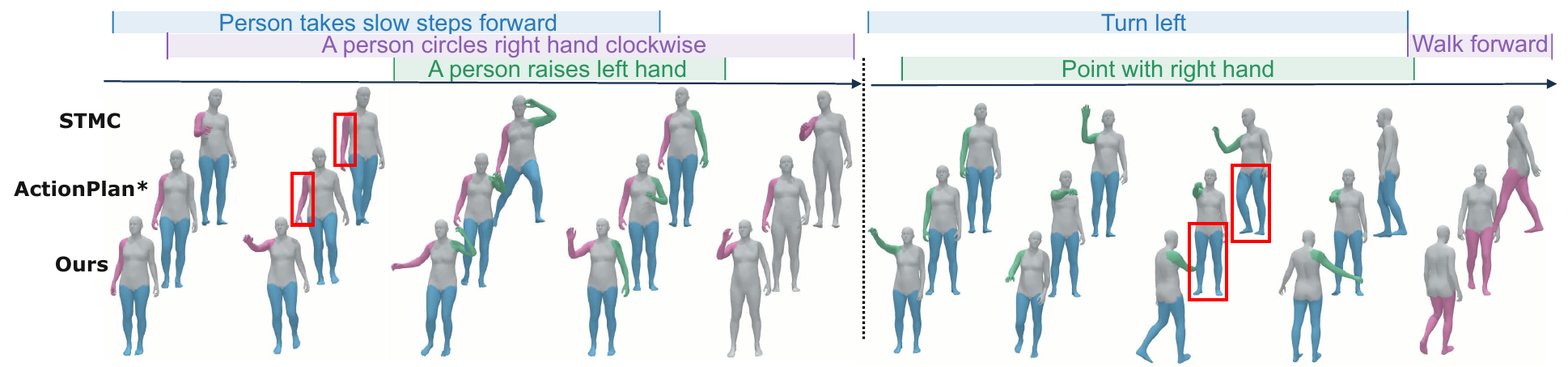}
    \caption{\textbf{Qualitative evaluation.}  We compare STMC, ActionPlan* and \method on \dataset (left) and MTT (right) dataset. Red box marks the failure part. \method is the only one not failing by ignoring instructions or confusing them.}
    
    \label{fig:qualitative}
\end{figure}

\noindent\textbf{Qualitative results.} We show qualitatives in Figure~\ref{fig:qualitative}, comparing with ActionPlan* and STMC, the best baselines. In presence of overlapping instructions, \textit{none of the baselines is able to correctly render the motion}. In particular, on the left, the instruction is fully ignored by ActionPlan*, while STMC renders it only in the very last frame, when there is no overlapping. On the right example, instead, ActionPlan ignores the instruction, while STMC mixes information in different labels and turn the human right rather than left. \method is instead able to correctly generate the desired motion, and to map each instruction into the correct interval. More examples are in Appendix~\ref{app:additional_qualitative} and also the supplementary video.

\noindent\textbf{User study.}
We conduct a user study to complement the evaluation in three aspects: text alignment, motion quality and temporal adherence. \method consistently outperforms other baselines. In particular, \method wins 66.2\% and 72.1\% on text-motion consistency against STMC and ActionPlan*, respectively.
More detailed results and the study design can be found in Appendix~\ref{app:user_study_detail}.

\paragraph{Autoencoder design.} Table~\ref{tab:autoencoder} positions our autoencoder design with published alternatives. We compare with 16- and 48-dimensional temporal autoencoders (TAE, 16-D and 48-D respectively)~\citep{xiao2025motionstreamer}, SALAD~\citep{hong2025salad}, and our part-structured autoencoder. We evaluate both reconstruction quality and downstream generation effectiveness.
All the 48-D autoencoders use 48-D body latents with 20-dimensional explicit root features (68 dimensions total). 
At the same 48-D body latent size, our part-aware autoencoder achieves comparable reconstruction FID on \dataset while reducing generation FID from 0.040 to 0.017 and improving retrieval on both datasets.
This result supports preserving anatomical structure in the features supplied to the denoising network. Finally, we show that removing our skeleton graph substantially degrades metrics (ours w/o sk.graph, Table~\ref{tab:autoencoder}), further justifying our design choices.
\begin{table}[!t]
    \centering
    \caption{\textbf{Autoencoder benchmarks.} Quantitative comparison of different autoencoder designs in reconstruction and generation in both HumanML3D and \dataset.}
    \label{tab:autoencoder}

    \small
    \setlength{\tabcolsep}{3pt}
    \renewcommand{\arraystretch}{1.05}

    \resizebox{\linewidth}{!}{%
    \begin{tabular}{@{}lcc*{11}{c}@{}}
        \toprule

        \multirow[c]{4}{*}{\textbf{Autoencoder}}
        & \multirow[c]{4}{*}{\textbf{Dims}}
& \multirow[c]{4}{*}{\shortstack{\textbf{Params}\\\textbf{(M)}}}        & \multicolumn{4}{c}{\textbf{\textit{HumanML3D}}}
        & \multicolumn{7}{c}{\textbf{\textit{\dataset}}}
        \\

        \cmidrule(lr){4-7}
        \cmidrule(lr){8-14}

        &
        &
        & 
        \multicolumn{2}{c}{{Reconstruction}}
        & \multicolumn{2}{c}{{Generation}}
        & \multicolumn{2}{c}{{Reconstruction}}
        & \multicolumn{3}{c}{{Generation}}
        & \multicolumn{2}{c}{{Temporal adherence}}
        \\

        \cmidrule(lr){4-5}
        \cmidrule(lr){6-7}
        \cmidrule(lr){8-9}
        \cmidrule(lr){10-12}
        \cmidrule(lr){13-14}

        &
        & 
        &
        FID $\downarrow$
        & MPJPE $\downarrow$
        & FID $\downarrow$
        & R@3$_{32}$ $\uparrow$
        & FID $\downarrow$
        & MPJPE $\downarrow$
        & FID $\downarrow$
        & FID+ $\downarrow$
        & R@3$_{32}$ $\uparrow$
        & AP $\uparrow$
        & AUC $\uparrow$
        \\

        \midrule

        TAE~\citep{xiao2025motionstreamer}
        & 16
        & 76.20
        & 0.0028
        & 21.9
        & 0.047
        & 0.833
        & 0.0020
        & 21.7
        & 0.051
        & 0.203
        & 0.533
        & 0.823
        & 0.755
        \\

        TAE~\citep{xiao2025motionstreamer}
        & 48
        & 79.42
        & \underline{0.0005}
        & \underline{3.19}
        & 0.049
        & 0.851
        & \textbf{0.0002}
        & \underline{3.77}
        & 0.040
        & \underline{0.197}
        & 0.576
        & \underline{0.827}
        & 0.762
        \\

        SALAD~\citep{hong2025salad}
        & 48
        & 0.51
        & 0.0034
        & 24.1
        & \underline{0.030}
        & \underline{0.858}
        & 0.0036
        & 27.7
        & \underline{0.023}
        & 0.208
        & 0.575
        & 0.820
        & 0.755
        \\

        Ours w/o sk.graph
        & 48
        & 4.80
        & 0.0012
        & 7.46
        & 0.038
        & 0.851
        & \underline{0.0007}
        & 8.27
        & 0.025
        & 0.209
        & \underline{0.600}
        & \textbf{0.832}
        & \underline{0.767}
        \\

        Ours
        & 48
        & 5.59
        & \textbf{0.0004}
        & \textbf{3.09}
        & \textbf{0.027}
        & \textbf{0.874}
        & \textbf{0.0002}
        & \textbf{3.58}
        & \textbf{0.017}
        & \textbf{0.192}
        & \textbf{0.603}
        & \textbf{0.832}
        & \textbf{0.772}
        \\

        \bottomrule
    \end{tabular}%
    }
\end{table}

\subsection{Ablation studies}
Table~\ref{tab:ablation_backbone} evaluates our architectural choices on \dataset, to assess streaming performance, and on HumanML3D, for evaluating loss of non-streaming capabilities. Further ablations on attention normalization, self-conditioning, and training-window sampling can be found in Appendix~\ref{app:training_strategies}.

\begin{table}[!t]
    \centering
    \caption{\textbf{Architectural ablations.} Ablation of temporal conditioning and part-aware denoising on HumanML3D and \dataset. Each variant removes the indicated component from \method. All our contributions together yield the best performance.}
    \label{tab:ablation_backbone}
    \small
    \setlength{\tabcolsep}{1.8pt}
    \resizebox{\textwidth}{!}{%
    \begin{tabular}{lcccccccccccc}
        \toprule
        \multirow{4}{*}{\textbf{Setup}} & \multicolumn{3}{c}{\textbf{\textit{HumanML3D}}} & \multicolumn{9}{c}{\textbf{\textit{\dataset}}}\\
        \cmidrule(lr){2-4}\cmidrule(lr){5-13}
        & \multirow{2}{*}{R@3$_{32}$ $\uparrow$} & \multirow{2}{*}{M2T $\uparrow$} & \multirow{2}{*}{FID $\downarrow$}
        & \multicolumn{2}{c}{{Sequence alignment}} & \multicolumn{2}{c}{{Body-part alignment}} & \multicolumn{3}{c}{{Motion quality}} & \multicolumn{2}{c}{{Temporal adherence}}\\
        \cmidrule(lr){5-6}\cmidrule(lr){7-8}\cmidrule(lr){9-11}\cmidrule(lr){12-13}
        &  &  &  & R@3$_{32}$ $\uparrow$ & M2T $\uparrow$ & R@3$_{32}$ $\uparrow$ & M2T $\uparrow$ & FID $\downarrow$ & FID+ $\downarrow$ & Trans $\downarrow$ & AP $\uparrow$ & AUC $\uparrow$\\
        \midrule
        Full model (Ours) & \textbf{0.874} & \textbf{0.769} & \textbf{0.027} & \textbf{0.603} & \textbf{0.653} & \textbf{0.374} & \textbf{0.553} & \textbf{0.017} & \textbf{0.192} & 1.31 & \textbf{0.832} & \textbf{0.772}\\
        \midrule
        RoTE only & 0.830 & 0.753 & 0.044 & 0.586 & 0.649 & 0.370 & \underline{0.552} & 0.028 & 0.197 & 1.28 & 0.820 & 0.752\\
        w/o RoTE & 0.820 & 0.748 & 0.061 & 0.572 & 0.648 & \underline{0.372} & 0.550 & 0.037 & 0.195 & \textbf{1.09} & 0.801 & 0.734\\
        \midrule
        w/o part-aware bias $B$ & \underline{0.850} & \underline{0.764} & \underline{0.038} & \underline{0.591} & \underline{0.650} & 0.357 & 0.547 & \underline{0.025} & \underline{0.193} & \underline{1.20} & \underline{0.830} & \underline{0.769}\\
        w/o part branch & \underline{0.850} & 0.763 & 0.042 & 0.588 & 0.649 & 0.350 & 0.545 & 0.030 & 0.198 & 1.26 & 0.829 & 0.767\\
        \bottomrule
    \end{tabular}%
    }
\end{table}

We ablate conditioning components in Table~\ref{tab:ablation_backbone} (top). We first remove the $M$ and $S$ masks and rely naively on RoTE only for multi-timeline conditioning (RoTE only). This reduces retrieval while increasing FID on both datasets, demonstrating the need of our attention masking for streaming multi-track motion generation. Removing RoTE (\textit{w/o RoTE}) leaves the masks intact but removes temporal sequence information from the encoded text. This has significant effects, with R@3 decreasing from \textbf{0.603} to 0.572 on \dataset. This shows that to benefit from our conditioning mechanism, we need appropriate encoding of the timeline arrival times and order.

We now ablate components of our part-aware denoising algorithm in Table~\ref{tab:ablation_backbone} (bottom). We first remove the bias $B$ from the multi-prompt conditioning (\textit{w/o part-aware bias $B$}). Most metrics decrease, showing that our representation exploits correctly the attention on multiple body parts, which $B$ enhances. We also propose a \textit{w/o part branch} setup that only uses the global state $\mathbf{g}$ to predict velocity. Note that $\mathbf{g}$ still receives information on the prompts with online multi-prompt conditioning. 
From the results, we considerably increase FID (\textbf{0.027}$\rightarrow$0.042 on HumanML3D, \textbf{0.017}$\rightarrow$0.030 on \dataset), showing that the task highly benefits from part-aware processing.

\subsection{Applications}
We propose some applications built on \method, to demonstrate the versatility of our method. 
In Figure~\ref{fig:teaser}(b), we show spatial conditioning, with a strategy like in~\citet{zhao2026ardy}. We can use the obtained model for position control on animation and robotics. We also deployed \method on a humanoid robot, executing motions in real time using SONIC~\citep{luo2026sonic} as motion tracker (Figure~\ref{fig:teaser}(c)).
Additional results and details are in Appendices~\ref{app:control_details} and~\ref{app:robot_execution}.

\section{Conclusion}
\label{sec:conclusion}

We introduced streaming multi-track timeline control, where independently timed instructions arrive during motion generation and may overlap. Our framework, TimelineControl, combines interval-aware text conditioning with part-aware denoising to coordinate concurrent actions. We also constructed TimelineMotion to support training and evaluation in this setting. Experiments show improved performance over the evaluated baselines on TimelineMotion and MTT. We further demonstrated spatial conditioning and execution of generated motions on a humanoid robot. Future work will explore online revision and cancellation of instructions, as well as resolving conflicting requests while preserving motion continuity.

\section*{Acknowledgements}
This work was in part supported by the Institute of Information \& communications Technology Planning \& Evaluation (IITP) grant funded by the Korean government (MSIT) (No. RS-2024-00457882, National AI Research Lab Project), and by the ANR project CorVis (ANR-21-CE230003-01).

\section*{AI use statement}
Generative AI tools assisted with manuscript polishing, code implementation and debugging, and body-part annotation of motion descriptions.
The annotation procedure and label review are described in Section~\ref{sec:data} and Appendix~\ref{app:data_construction}.

\section*{Reproducibility statement}
The model architecture and conditioning mechanisms are given in Sections~\ref{sec:autoencoder}--\ref{sec:part-aware-denoising}, with implementation,
training-objective, and streaming-inference details in
Appendix~\ref{app:implementation}.
The \dataset construction procedure is given in
Section~\ref{sec:data} and Appendix~\ref{app:data_construction},
including the annotation, compatibility, transition-refinement, and filtering
rules.
The experimental setup and evaluation metrics are given in Section~\ref{subsec:setup} and Appendix~\ref{app:evaluation_protocols}.
Additional training strategies and ablations are provided in Appendix~\ref{app:training_strategies}.
Our code, data and models will be released.

\bibliography{iclr2027_conference}
\bibliographystyle{iclr2027_conference}
\clearpage

\appendix

\section{\dataset Construction Details}
\label{app:data_construction}

\begin{figure}[t]
    \centering
    \includegraphics[width=\textwidth]{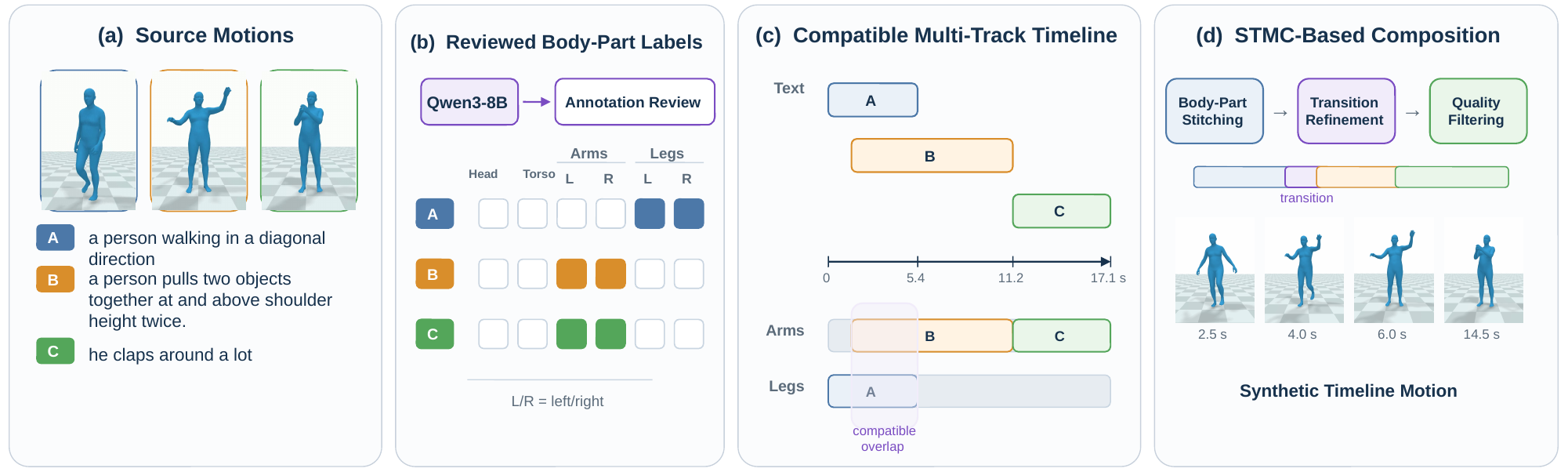}
    \caption{Construction of \dataset. (a) HumanML3D source motions and captions. (b) Body-part annotations. (c) Compatible instructions arranged into multi-track timelines. (d) STMC-based body-part composition, transition refinement, and quality filtering.}
    \label{fig:data_construction}
\end{figure}

Figure~\ref{fig:data_construction} summarizes the TimelineMotion construction pipeline; below, we provide the annotation and filtering details omitted from Section~\ref{sec:data}.

\subsection{Body-Part Annotation}

Qwen3-8B~\citep{qwen3} annotates captions from HumanML3D dataset with body-part labels and motion attributes. The primary labels specify the anatomical regions used for motion composition, part supervision, and body-part evaluation. Primary labels that refer to the upper, lower, or full body are expanded into the corresponding anatomical regions when explicit regional labels are absent.

The annotation prompt below uses \texttt{\{text\}} for the input motion caption.

{\footnotesize
\begin{verbatim}
You are an expert motion-text parser.
Parse the input action text into one JSON
object with only these fields:
- primary_parts: list[str] from
  ["torso", "head", "left_arm",
   "right_arm", "left_leg", "right_leg",
   "upper_body", "lower_body", "full_body"]
- secondary_parts: list[str] with same
  allowed labels
- coarse_parts: list[str] from
  ["upper_body", "lower_body", "full_body"]
- locomotion: bool
- stance_change: bool
- whole_body: bool
- dynamic_level: one of
  ["low", "medium", "high", "unknown"]
- stance_type: one of
  ["standing", "sitting", "crouching",
   "lying", "unknown"]

Rules:
- Do not output explanation.
- Output JSON only.
- If text mentions hand/arm gestures
  (wave, clap, throw, catch, point, raise
  hands, etc.), include explicit arm labels
  in primary_parts.
- If side is unspecified for arm gestures,
  include both "left_arm" and "right_arm".
- If text mentions leg/foot/lower-body
  actions (kick, step, walk, run, jump,
  squat, etc.), include explicit leg labels
  in primary_parts.
- If side is unspecified for leg/foot
  actions, include both "left_leg" and
  "right_leg".
- Do not use only coarse labels when
  explicit limbs are inferable from text
  (same principle for leg/foot actions).

Input text:
{text}
\end{verbatim}
}

\subsection{Timeline Construction}

We randomly sample source clips from the same HumanML3D split and consider sequential or overlapping temporal arrangements. For overlapping clips, we exclude pairs that require the same body regions, contain conflicting root motion or locomotion, involve incompatible poses or large stance changes, or contain full-body actions. The primary body-part labels determine the source motion assigned to each anatomical region during STMC composition, as described in Section~\ref{sec:data}.

We refine transition and overlap regions with interpolation to improve continuity while retaining the surrounding source motion.
Following InterPose~\citep{zhang2026interpose}, refinement uses spherical interpolation for joint rotations and cubic Hermite interpolation where root trajectories require smoothing.
We additionally identify composition artifacts around timeline boundaries from decoded joint motion. We first express joint positions relative to the pelvis and compute the magnitude of their temporal second differences, averaged over joints. For each composition boundary, we take the maximum value within two frames of the boundary as the boundary peak. As a local reference, we compute the 95th percentile within 12 frames on either side of the boundary, excluding the three frames closest to the boundary on each side. We reject a sequence when the boundary peak is at least $0.03$, exceeds the local reference by at least $0.025$, and is at least $1.5\times$ the local reference. 
We also remove sequences containing non-finite motion values.

\section{Objectives and Implementation Details}
\label{app:objectives}

\subsection{Causal Part-Structured Autoencoder}
\label{app:autoencoder}

\begin{figure}[htbp]
    \centering
    \includegraphics[width=\textwidth]{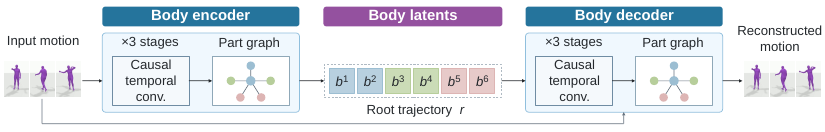}
    \caption{Causal part-structured autoencoder. Causal temporal convolutions and skeleton-graph message passing encode and reconstruct six anatomical body regions. The root trajectory bypasses the body encoder and conditions the decoder.}
    \label{fig:part_tae}
\end{figure}

Figure~\ref{fig:part_tae} illustrates the architecture described in Section~\ref{sec:autoencoder}.
Body features comprise 21 non-root joint positions, 22 joint rotations in 6D, joint velocities, and four foot-contact indicators.
Positions retain world-space height but are horizontally centered at the pelvis. Rotations and velocities use world coordinates.
The torso includes root rotation and velocity, and each leg includes its foot-contact indicators.
Following MotionStreamer~\citep{xiao2025motionstreamer}, the autoencoder objective is
\begin{equation}
\begin{split}
    \mathcal{L}_{\mathrm{AE}}
    ={}&\mathcal{L}_{\mathrm{rec}}+\mathcal{L}_{\mathrm{KL}}
       +\mathcal{L}_{\mathrm{acc}}.
\end{split}
    \label{eq:ae_loss}
\end{equation}
Here, \(\mathcal{L}_{\mathrm{rec}}\) is the Gaussian negative log-likelihood reconstruction loss and \(\mathcal{L}_{\mathrm{KL}}\) regularizes the latent posterior. The temporal loss \(\mathcal{L}_{\mathrm{acc}}\) matches second-order differences.

\subsection{Denoiser Architecture and Training Settings}
\label{app:implementation}

The global branch has four stages of four Transformer layers, with width 1024, feed-forward width 4096, and 16 attention heads.
The input token encoder has one attention layer with width 128 and four heads.
The classifier-predicted attention bias uses \(\lambda_{\mathrm{r}}=0.25\).
The classifier takes projected prompt embeddings as input.
Body-part tokens use centered relevance probabilities to form $B^k$, while the root uses $B^0=0$.
We stop gradients at the classifier input and through the probabilities used to form $B^k$ in Section~\ref{sec:part-aware-denoising}.
The generator learning rate is 1e-4, with weight decay 0.01. HumanML3D examples are capped at 78 latents with a latent rate of 7.5\,fps.
RoTE uses frequencies \(\omega_j=10000^{-(j-1)/D_{\mathrm{rot}}}\) for \(j=1,\ldots,D_{\mathrm{rot}}\), where \(D_{\mathrm{rot}}\) is the number of rotary pairs.
A common timestamp scale \(\gamma=4\) is used for motion queries and interval centers and radii.

\subsection{Backbone Objectives and details}

Following ActionPlan~\citep{nazarenus2026actionplan}, we construct action-plan targets from atomic action annotations in BABEL~\citep{punnakkal2021babel}, aligned to the source motion clips and transferred to the composed timelines.
We average the CLIP embeddings of distinct active action labels at each frame, temporally smooth the features, and compress them with a frozen autoencoder.
We apply the interpolation and velocity target in Eq.~\eqref{eq:rectified-flow} to both streams.
Each motion latent uses a noise level \(t_n^{\mathrm m}\), while the action-plan stream shares one level \(t^{\mathrm p}\) per training example.

The weighted flow loss is computed over scalar targets in the minibatch and normalized by the total weight jointly across the motion and action-plan streams.
Valid motion targets use importance weights that compensate for their nonuniform noise-level distribution. Valid action-plan targets have unit weight.
The shared action-plan noise level is sampled uniformly.
Motion supervision excludes the fixed prefix, while available action-plan targets at prefix positions remain supervised.
Padding, positions beyond the supervised horizon, and targets replaced by detached self-conditioned predictions have zero weight.

The main generation objective is
\begin{equation}
\begin{split}
    \mathcal{L}_{\mathrm{gen}}={}&\mathcal{L}_{\mathrm{flow}}
        +0.2\mathcal{L}_{\mathrm{part}}
        +0.02\mathcal{L}_{\mathrm{cls}}
         +\mathcal{L}_{\mathrm{root}}.
\end{split}
    \label{eq:full_loss}
\end{equation}
Here, \(\mathcal{L}_{\mathrm{part}}\) is the mean squared error of the auxiliary body-latent flow-velocity prediction.
Let $Y\in\{0,1\}^{J\times6}$ denote the annotated body-part labels, with the same instruction and part ordering as $\Pi$.
\(\mathcal{L}_{\mathrm{cls}}\) uses binary cross-entropy with $Y$.
The root loss \(\mathcal{L}_{\mathrm{root}}\) applies smooth L1 supervision to frame-to-frame changes in horizontal position and heading.

\subsection{Streaming Inference}
\label{app:streaming_inference}

Following ActionPlan~\citep{nazarenus2026actionplan}, each sampling window retains up to \(P\) generated motion latents as a fixed prefix and initializes future motion and all action-plan latents with Gaussian noise.
The action-plan stream is first denoised jointly with the first motion token, while later motion positions remain at higher noise levels.
Once the action plan is complete, its latents remain fixed and motion is denoised progressively.
Prompt attention follows Eq.~\eqref{eq:prompt_mask} throughout sampling.
Completed motion is committed to the output and extends the fixed prefix.
When a new instruction arrives, committed motion is preserved, while uncommitted predictions are discarded and regenerated using the new instruction together with earlier instructions that remain active. 
For open-ended interactive commands without a specified end time, we use a 5-second default interval, which can be extended successively for continuous execution. We provide instructions in seconds and then map them to the window interval that the network is operating on.

\section{Evaluation Protocols and Metrics}
\label{app:evaluation_protocols}

\subsection{Baseline Text Conditioning}
\label{app:baseline_conditioning}

For baselines that accept only a single caption, we split the inference timeline at instruction start and end times.
Using the notation of Section~\ref{sec:preliminaries}, instruction $e_i=(\ell_i,[s_i,u_i))$ contributes its text exactly when $s_i\leq\tau<u_i$.
Within each resulting segment, a single active prompt is passed unchanged, and multiple active prompts are joined with \emph{while}.
The caption is recomputed at each boundary, removing instructions that have ended and adding instructions that have just arrived.
For example, prompts A on $[0,5)$ and B on $[3,8)$ produce A on $[0,3)$, A \emph{while} B on $[3,5)$, and B on $[5,8)$.
Future prompts are withheld until their annotated start times, and caption updates retain the already emitted motion through the baseline's native continuation mechanism.
This inference procedure is separate from the window-level caption construction used during retraining.

\subsection{Evaluation Metrics}
\label{app:evaluation_metrics}

\paragraph{Sequence alignment.}
We use TMR~\citep{petrovich2023tmr} to measure motion-to-text R-Precision at ranks 1, 2, and 3, motion-to-text similarity (M2T), and similarity to the corresponding source motion (M2M), with both similarities rescaled to $[0,1]$.
On timeline benchmarks, we evaluate text-motion alignment within each prompt's annotated interval.
Overlapping prompts are scored separately over their respective intervals, so shared frames contribute to the evaluation of each prompt.
On HumanML3D, alignment is evaluated over complete motions.

\paragraph{Body-part alignment.}
For body-part alignment, we follow FrankenMotion's released protocol~\citep{li2026frankenmotion}, scoring each prompt's full-body motion crop against its original text with the evaluators for its primary body-part labels.
The same evaluators measure M2M between each generated interval and its corresponding source motion.
We report R@3, M2T, and M2M averaged equally over the six body regions on \dataset and MTT.

\paragraph{Motion quality.}
FID measures distributional similarity using whole motion sequences. FID+ compares five crops of up to five seconds per generated timeline with ground truth source motions, following STMC~\citep{petrovich2024stmc}.
We report both on \dataset, FID on HumanML3D, and FID+ on MTT.
Because MTT has no complete reference timelines, we omit its whole-motion FID and reference-motion temporal-adherence scores.
Transition distance (Trans) measures root-relative joint displacement at instruction boundaries~\citep{athanasiou2022teach,petrovich2024stmc}.

\paragraph{Reconstruction.}
FID measures distributional similarity between reconstructed and reference motions.
MPJPE measures the Euclidean distance between corresponding reconstructed and reference joint positions, averaged over joints and all valid frames and reported in millimeters.

\paragraph{Temporal adherence.}
For temporal adherence, we measure whether each instruction is reflected within its requested interval using frozen whole-body TMR embeddings.
For every instruction $e_i=(\ell_i,[s_i,u_i))$, we score its text against 2-second motion windows sampled every 0.2 seconds across the complete timeline.
A window is positive if its center lies in $[s_i,u_i)$ and negative otherwise.
We report non-interpolated average precision (AP) and the area under the receiver operating characteristic curve (ROC-AUC). AP is computed from the ranking induced by TMR similarity scores, without selecting a fixed similarity threshold.
We score overlapping instructions independently and average each metric equally over instructions with both positive and negative windows.
Temporal adherence is not applicable to HumanML3D because its captions span complete motions.

\subsection{Temporal Adherence Validation}
\label{app:temporal_validation}

To assess sensitivity to temporal misalignment, we keep each motion and its text-motion similarity scores fixed and shift the requested instruction interval earlier or later by 0.5, 1, or 2 seconds.
We evaluate GT instruction intervals from the \dataset test set.
Interval lengths and scoring windows are unchanged across shifts.
Table~\ref{tab:temporal_adherence_validation} shows that mean AP and ROC-AUC are highest at the original intervals for GT.
On average, shifting intervals in either direction lowers both metrics, with larger shifts producing larger decreases.
These results support using Temporal AP and ROC-AUC to evaluate alignment between motion content and requested instruction intervals.

\begin{table}[!t]
    \centering
    \caption{Sensitivity of temporal adherence metrics under shifts of the requested instruction intervals on \dataset. Best and second-best scores are bold and underlined.}
    \label{tab:temporal_adherence_validation}
    \begin{tabular}{lcc}
        \toprule
        Interval shift & AP $\uparrow$ & ROC-AUC $\uparrow$ \\
        \midrule
        - 2 s & 0.440 & 0.564\\
        - 1 s & 0.580 & 0.675\\
        - 0.5 s & \underline{0.630} & 0.712 \\
        0 & \textbf{0.647} & \textbf{0.728}\\
        + 0.5 s & 0.622 & \underline{0.713}\\
        + 1 s & 0.556 & 0.671\\
        + 2 s & 0.417 & 0.568\\
        \bottomrule
    \end{tabular}
\end{table}

\subsection{Runtime Measurement}
\label{app:runtime}

Runtime is evaluated on an NVIDIA RTX 6000 Ada at batch size one in FP32 with a sampling window of 78 latents.
After five warm-up runs, we report the mean latency to the first clean motion token and the mean interval between subsequent clean tokens over 100 sequences. Each token represents four motion frames.

\subsection{User Study Details}
\label{app:user_study_detail}

\begin{figure}[htbp]
    \centering
    \includegraphics[width=\textwidth]{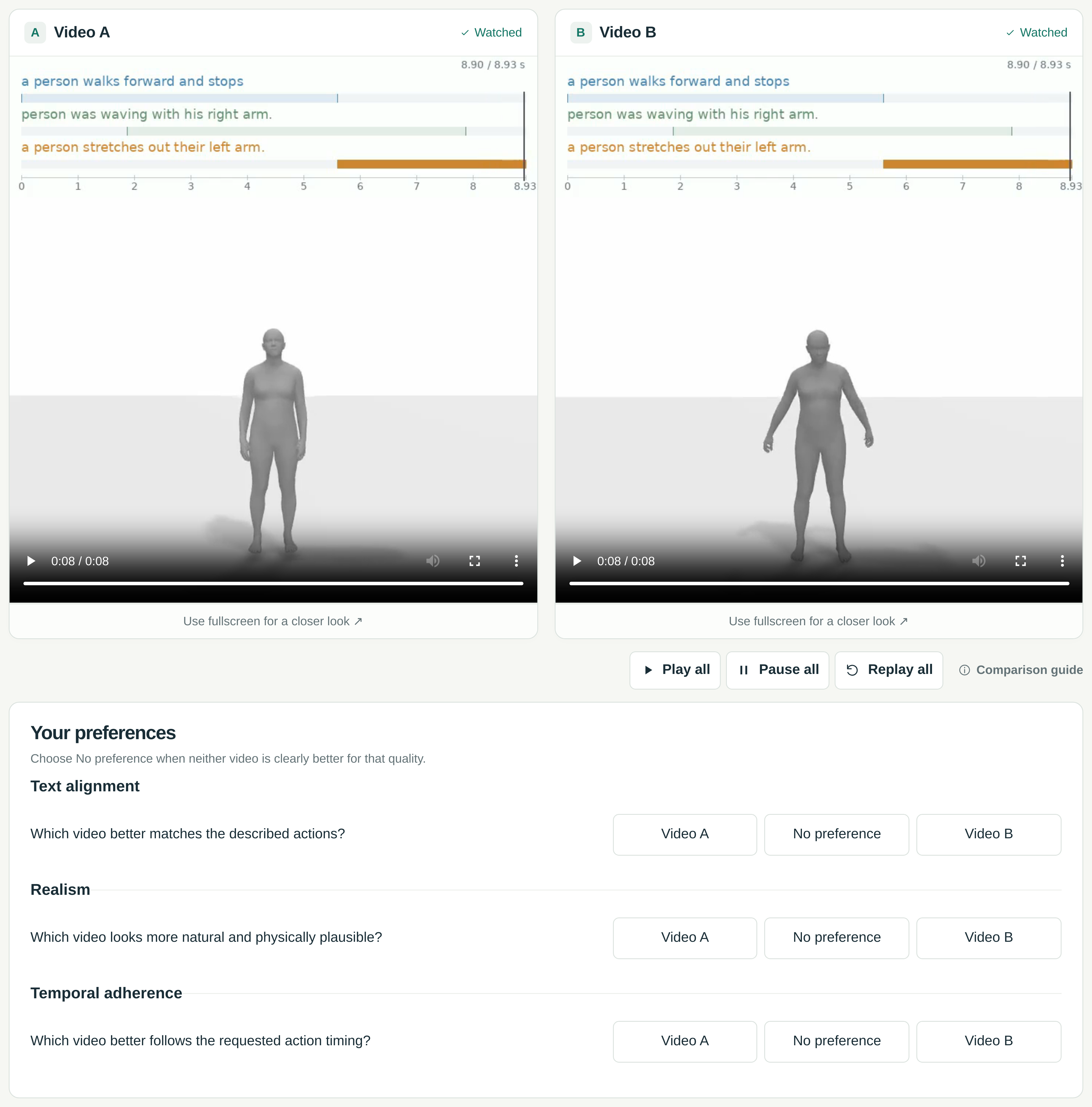}
    \caption{User study interface. We present timeline with videos generated by \method and one of the baselines. Users are asked to compare the two videos in terms of text-motion alignment, motion quality, and temporal adherence.}
    \label{fig:user_study_web}
\end{figure}

\begin{figure}[t]
    \centering
    \includegraphics[width=\textwidth]{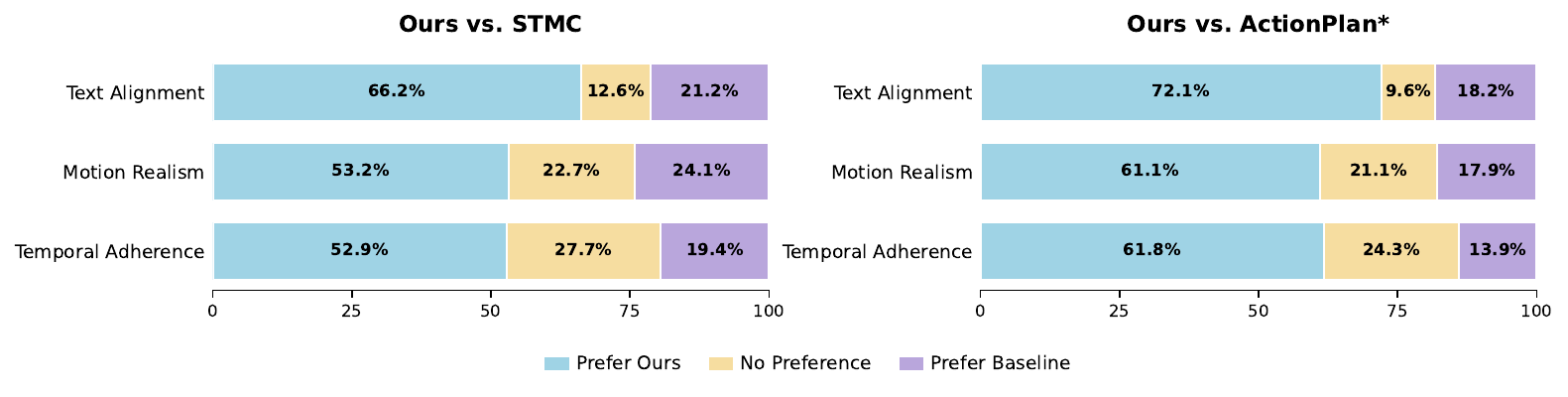}
    \caption{User study. Comparison for timeline human motion videos generated by \method, STMC and ActionPlan*.}
    \label{fig:user_study}
\end{figure}

We randomly generated 20 multi-track human motion videos with STMC~\citep{petrovich2024stmc}, ActionPlan*~\citep{nazarenus2026actionplan} and \method. As shown in Figure~\ref{fig:user_study_web}, for each question, we put the videos from \method and one of the baseline methods in random order to form 40 questions in total. For each timeline input, 18 participants compare each pair with respect to text alignment, motion realism, and temporal adherence. 
As presented in Figure~\ref{fig:user_study}, \method consistently outperforms other baselines in all aspects.

\section{Training Strategies and Additional Experiments}
\label{app:training_strategies}

\subsection{Self-Conditioned Training}

During streaming inference, previously generated latents become context for subsequent predictions.
Following MotionStreamer~\citep{xiao2025motionstreamer}, we train with detached model predictions as context.
For examples selected for replacement, a first forward pass estimates clean motion and action-plan latents without gradient propagation.
We insert the selected predictions into the second-pass input as clean latents and set their noise levels to zero.
Gradients are computed only through the second pass. Replaced positions receive zero flow-loss weight in the corresponding stream, while the remaining valid targets retain supervision.

Table~\ref{tab:ablation_self_conditioning} compares the full model with its variant without self-conditioned training.
Removing self-conditioned training lowers retrieval and increases FID on both datasets, and increases boundary displacement on \dataset.
Temporal AP is unchanged at the reported precision, while ROC-AUC decreases slightly.

\begin{table}[!t]
    \centering
    \caption{Effect of self-conditioned training on motion generation on HumanML3D and \dataset. Best results are bold.}
    \label{tab:ablation_self_conditioning}
    \small
    \setlength{\tabcolsep}{1.8pt}
    \resizebox{\textwidth}{!}{%
    \begin{tabular}{lcccccccccccc}
        \toprule
        \multirow{3}{*}{Variant} & \multicolumn{3}{c}{HumanML3D} & \multicolumn{9}{c}{\dataset}\\
        \cmidrule(lr){2-4}\cmidrule(lr){5-13}
        & \multirow{2}{*}{R@3$_{32}$ $\uparrow$} & \multirow{2}{*}{M2T $\uparrow$} & \multirow{2}{*}{FID $\downarrow$}
        & \multicolumn{2}{c}{Sequence alignment} & \multicolumn{2}{c}{Body-part alignment} & \multicolumn{3}{c}{Motion quality} & \multicolumn{2}{c}{Temporal adherence}\\
        \cmidrule(lr){5-6}\cmidrule(lr){7-8}\cmidrule(lr){9-11}\cmidrule(lr){12-13}
        &  &  &  & R@3$_{32}$ $\uparrow$ & M2T $\uparrow$ & R@3$_{32}$ $\uparrow$ & M2T $\uparrow$ & FID $\downarrow$ & FID+ $\downarrow$ & Trans (cm) $\downarrow$ & AP $\uparrow$ & ROC-AUC $\uparrow$\\
        \midrule
        Full model (Ours) & \textbf{0.874} & \textbf{0.769} & \textbf{0.027} & \textbf{0.603} & \textbf{0.653} & \textbf{0.374} & \textbf{0.553} & \textbf{0.017} & \textbf{0.192} & \textbf{1.31} & \textbf{0.832} & \textbf{0.772}\\
        w/o self-conditioned training & 0.862 & 0.767 & 0.035 & 0.594 & 0.650 & 0.365 & 0.551 & 0.024 & 0.198 & 1.35 & \textbf{0.832} & 0.770\\
        \bottomrule
    \end{tabular}%
    }
\end{table}

\subsection{Training-Window Strategies}

For examples from \dataset, we choose the prediction start either uniformly across the sequence (\emph{Random start}) or at the start of a text instruction (\emph{Instruction start}), with equal probability.
Random-start crops cover different stages of motion, while instruction-start crops emphasize responses to newly received instructions.
Each crop includes up to \(P\) preceding clean motion latents as context. Action-plan latents remain prediction targets.
Table~\ref{tab:ablation_sampling} compares the full training-window policy with Random start and Instruction start variants.
The mixture achieves the highest sequence retrieval and lowest FID on both datasets, and the highest Temporal AP and ROC-AUC on \dataset.
The Random start baseline has the lowest FID+, while the Instruction start baseline has the lowest boundary displacement.

\begin{table}[!t]
    \centering
    \caption{Comparison of training-window strategies on HumanML3D and \dataset.}
    \label{tab:ablation_sampling}
    \small
    \setlength{\tabcolsep}{1.8pt}
    \resizebox{\textwidth}{!}{%
    \begin{tabular}{lcccccccccccc}
        \toprule
        \multirow{3}{*}{Training-window strategy} & \multicolumn{3}{c}{HumanML3D} & \multicolumn{9}{c}{\dataset}\\
        \cmidrule(lr){2-4}\cmidrule(lr){5-13}
        & \multirow{2}{*}{R@3$_{32}$ $\uparrow$} & \multirow{2}{*}{M2T $\uparrow$} & \multirow{2}{*}{FID $\downarrow$}
        & \multicolumn{2}{c}{Sequence alignment} & \multicolumn{2}{c}{Body-part alignment} & \multicolumn{3}{c}{Motion quality} & \multicolumn{2}{c}{Temporal adherence}\\
        \cmidrule(lr){5-6}\cmidrule(lr){7-8}\cmidrule(lr){9-11}\cmidrule(lr){12-13}
        &  &  &  & R@3$_{32}$ $\uparrow$ & M2T $\uparrow$ & R@3$_{32}$ $\uparrow$ & M2T $\uparrow$ & FID $\downarrow$ & FID+ $\downarrow$ & Trans (cm) $\downarrow$ & AP $\uparrow$ & ROC-AUC $\uparrow$\\
        \midrule
        Random start & 0.859 & 0.762 & 0.035 & \underline{0.597} & 0.651 & \textbf{0.374} & \textbf{0.553} & 0.027 & \textbf{0.187} & 1.48 & 0.829 & 0.765\\
        Mixture (Ours) & \textbf{0.874} & \textbf{0.769} & \textbf{0.027} & \textbf{0.603} & \textbf{0.653} & \textbf{0.374} & \textbf{0.553} & \textbf{0.017} & \underline{0.192} & \underline{1.31} & \textbf{0.832} & \textbf{0.772}\\
        Instruction start & \underline{0.866} & \underline{0.764} & \underline{0.029} & 0.596 & \underline{0.652} & \underline{0.371} & \underline{0.551} & \underline{0.019} & 0.196 & \textbf{1.28} & \underline{0.831} & \underline{0.768}\\
        \bottomrule
    \end{tabular}%
    }
\end{table}

\subsection{Attention Normalization}
\label{app:attention_normalization}

Table~\ref{tab:ablation_attention} compares sigmoid weighting with softmax normalization in text attention.
The softmax variant replaces the weighting in both the part-aware and global branches, while retaining the temporal interval encodings, visibility masks, and part-aware biases.
Sigmoid weighting yields higher sequence retrieval and lower whole-motion FID on both datasets, and higher Temporal AP and ROC-AUC on \dataset.
Softmax yields slightly lower FID+ and boundary displacement on \dataset.

\begin{table}[!t]
    \centering
    \caption{Comparison of sigmoid weighting and softmax normalization in multi-track attention on HumanML3D and \dataset.}
    \label{tab:ablation_attention}
    \small
    \setlength{\tabcolsep}{1.8pt}
    \resizebox{\textwidth}{!}{%
    \begin{tabular}{lcccccccccccc}
        \toprule
        \multirow{3}{*}{Variant} & \multicolumn{3}{c}{HumanML3D} & \multicolumn{9}{c}{\dataset}\\
        \cmidrule(lr){2-4}\cmidrule(lr){5-13}
        & \multirow{2}{*}{R@3$_{32}$ $\uparrow$} & \multirow{2}{*}{M2T $\uparrow$} & \multirow{2}{*}{FID $\downarrow$}
        & \multicolumn{2}{c}{Sequence alignment} & \multicolumn{2}{c}{Body-part alignment} & \multicolumn{3}{c}{Motion quality} & \multicolumn{2}{c}{Temporal adherence}\\
        \cmidrule(lr){5-6}\cmidrule(lr){7-8}\cmidrule(lr){9-11}\cmidrule(lr){12-13}
        &  &  &  & R@3$_{32}$ $\uparrow$ & M2T $\uparrow$ & R@3$_{32}$ $\uparrow$ & M2T $\uparrow$ & FID $\downarrow$ & FID+ $\downarrow$ & Trans (cm) $\downarrow$ & AP $\uparrow$ & ROC-AUC $\uparrow$\\
        \midrule
        Sigmoid (Ours) & \textbf{0.874} & \textbf{0.769} & \textbf{0.027} & \textbf{0.603} & \textbf{0.653} & \textbf{0.374} & \textbf{0.553} & \textbf{0.017} & 0.192 & 1.31 & \textbf{0.832} & \textbf{0.772}\\
        Softmax & 0.847 & 0.754 & 0.041 & 0.580 & 0.644 & \textbf{0.374} & 0.551 & 0.027 & \textbf{0.190} & \textbf{1.23} & 0.825 & 0.757\\
        \bottomrule
    \end{tabular}%
    }
\end{table}

\subsection{Multi-Text Conditioning in ActionPlan}
\label{app:actionplan_multitext}

Table~\ref{tab:ablation_actionplan_multitext} compares ActionPlan, its jointly retrained variant, and a variant with multi-text conditioning.
The multi-text variant encodes each instruction separately and masks its text cross-attention outside the annotated interval, while retaining the backbone depth and width and the same motion autoencoder.
It uses softmax text attention without our temporal interval encoding or part-aware branches.
Relative to the jointly retrained single-caption baseline, multi-text conditioning improves sequence retrieval and temporal adherence on \dataset.
On \dataset, our model further improves sequence and body-part alignment, FID, and temporal adherence, while both models have the same boundary displacement at the reported precision.
These results show that multi-text conditioning improves ActionPlan, but does not account for the full gain of our method.

\begin{table}[t]
    \centering
    \caption{Comparison of joint retraining and multi-text conditioning variants of ActionPlan with \method on HumanML3D and \dataset. $^{*}$ denotes joint retraining on both datasets. Multi-text uses separately encoded instructions with an interval visibility mask. Best and second-best results are bold and underlined.}
    \label{tab:ablation_actionplan_multitext}
    \small
    \setlength{\tabcolsep}{1.8pt}
    \resizebox{\textwidth}{!}{%
    \begin{tabular}{lcccccccccccc}
        \toprule
        \multirow{3}{*}{Variant} & \multicolumn{3}{c}{HumanML3D} & \multicolumn{9}{c}{\dataset}\\
        \cmidrule(lr){2-4}\cmidrule(lr){5-13}
        & \multirow{2}{*}{R@3$_{32}$ $\uparrow$} & \multirow{2}{*}{M2T $\uparrow$} & \multirow{2}{*}{FID $\downarrow$}
        & \multicolumn{2}{c}{Sequence alignment} & \multicolumn{2}{c}{Body-part alignment} & \multicolumn{3}{c}{Motion quality} & \multicolumn{2}{c}{Temporal adherence}\\
        \cmidrule(lr){5-6}\cmidrule(lr){7-8}\cmidrule(lr){9-11}\cmidrule(lr){12-13}
        &  &  &  & R@3$_{32}$ $\uparrow$ & M2T $\uparrow$ & R@3$_{32}$ $\uparrow$ & M2T $\uparrow$ & FID $\downarrow$ & FID+ $\downarrow$ & Trans (cm) $\downarrow$ & AP $\uparrow$ & ROC-AUC $\uparrow$\\
        \midrule
        ActionPlan & 0.829 & 0.724 & \textbf{0.022} & 0.405 & 0.585 & 0.303 & \underline{0.547} & 0.274 & 0.330 & 2.16 & 0.733 & 0.653\\
        ActionPlan$^{*}$ & 0.830 & \underline{0.734} & 0.044 & 0.502 & 0.622 & 0.347 & 0.546 & 0.055 & 0.209 & \textbf{0.92} & 0.784 & 0.711\\
        ActionPlan$^{*}$ + multi-text & \underline{0.845} & 0.731 & \underline{0.027} & \underline{0.566} & \underline{0.640} & \underline{0.352} & \underline{0.547} & \underline{0.036} & \underline{0.198} & \underline{1.31} & \underline{0.805} & \underline{0.745}\\
        \method & \textbf{0.874} & \textbf{0.769} & \underline{0.027} & \textbf{0.603} & \textbf{0.653} & \textbf{0.374} & \textbf{0.553} & \textbf{0.017} & \textbf{0.192} & \underline{1.31} & \textbf{0.832} & \textbf{0.772}\\
        \bottomrule
    \end{tabular}%
    }
\end{table}

\subsection{Ablation of Block-Causal Self-Attention}
\label{app:causal_mask}

We replace block-causal motion self-attention with a standard causal mask (\emph{w/o block-causal self-attention}).
Each motion token can attend only to itself and preceding tokens, removing bidirectional interactions within the jointly denoised block.
The online instruction-arrival policy and text-conditioning mechanism are retained.
Table~\ref{tab:ablation_causal_mask} shows that this replacement lowers sequence R@3 on \dataset from 0.603 to 0.586 and body-part R@3 from 0.374 to 0.364.
Temporal AP and ROC-AUC decrease from 0.832 to 0.826 and from 0.772 to 0.767, respectively.
These results suggest that within-block bidirectional interactions are beneficial for timeline alignment, temporal adherence, and distributional motion quality on \dataset.

\begin{table}[!t]
    \centering
    \caption{Comparison of block-causal and standard causal self-attention on HumanML3D and \dataset.}
    \label{tab:ablation_causal_mask}
    \small
    \setlength{\tabcolsep}{1.8pt}
    \resizebox{\textwidth}{!}{
    \begin{tabular}{lcccccccccccc}
        \toprule
        \multirow{3}{*}{Variant} & \multicolumn{3}{c}{HumanML3D} & \multicolumn{9}{c}{\dataset}\\
        \cmidrule(lr){2-4}\cmidrule(lr){5-13}
        & \multirow{2}{*}{R@3$_{32}$ $\uparrow$} & \multirow{2}{*}{M2T $\uparrow$} & \multirow{2}{*}{FID $\downarrow$}
        & \multicolumn{2}{c}{Sequence alignment} & \multicolumn{2}{c}{Body-part alignment} & \multicolumn{3}{c}{Motion quality} & \multicolumn{2}{c}{Temporal adherence}\\
        \cmidrule(lr){5-6}\cmidrule(lr){7-8}\cmidrule(lr){9-11}\cmidrule(lr){12-13}
        &  &  &  & R@3$_{32}$ $\uparrow$ & M2T $\uparrow$ & R@3$_{32}$ $\uparrow$ & M2T $\uparrow$ & FID $\downarrow$ & FID+ $\downarrow$ & Trans (cm) $\downarrow$ & AP $\uparrow$ & ROC-AUC $\uparrow$\\
        \midrule
        Full model & \textbf{0.874} & 0.769 & 0.027 & \textbf{0.603} & \textbf{0.653} & \textbf{0.374} & \textbf{0.553} & \textbf{0.017} & \textbf{0.192} & 1.31 & \textbf{0.832} & \textbf{0.772}\\
        w/o block-causal self-attention & 0.870 & \textbf{0.770} & \textbf{0.023} & 0.586 & 0.649 & 0.364 & \textbf{0.553} & 0.019 & 0.212 & \textbf{1.22} & 0.826 & 0.767\\
        \bottomrule
    \end{tabular}
    }
\end{table}

\subsection{Ablation of Online Multi-Track Attention}
\label{app:online_multitrack_attention}

We ablate separate instruction conditioning by merging the captions of currently active instructions into a single caption (\emph{w/o online multi-track attention}).
The merged caption is updated at instruction start and end boundaries during training and inference, and future instructions remain unavailable before arrival.
The model therefore conditions on one combined description in each interval rather than attending independently to concurrent instructions.
Table~\ref{tab:ablation_online_multitrack_attention} shows that sequence R@3 on \dataset decreases from 0.603 to 0.552 and whole-motion FID increases from 0.017 to 0.046.
The performance on HumanML3D also drops.
The comparison supports separate online instruction conditioning for semantic alignment, temporal adherence, and whole-motion distributional quality.

\begin{table}[!t]
    \centering
    \caption{Effect of online multi-track attention on HumanML3D and \dataset.}
    \label{tab:ablation_online_multitrack_attention}
    \small
    \setlength{\tabcolsep}{1.8pt}
    \resizebox{\textwidth}{!}{
    \begin{tabular}{lcccccccccccc}
        \toprule
        \multirow{3}{*}{Variant} & \multicolumn{3}{c}{HumanML3D} & \multicolumn{9}{c}{\dataset}\\
        \cmidrule(lr){2-4}\cmidrule(lr){5-13}
        & \multirow{2}{*}{R@3$_{32}$ $\uparrow$} & \multirow{2}{*}{M2T $\uparrow$} & \multirow{2}{*}{FID $\downarrow$}
        & \multicolumn{2}{c}{Sequence alignment} & \multicolumn{2}{c}{Body-part alignment} & \multicolumn{3}{c}{Motion quality} & \multicolumn{2}{c}{Temporal adherence}\\
        \cmidrule(lr){5-6}\cmidrule(lr){7-8}\cmidrule(lr){9-11}\cmidrule(lr){12-13}
        &  &  &  & R@3$_{32}$ $\uparrow$ & M2T $\uparrow$ & R@3$_{32}$ $\uparrow$ & M2T $\uparrow$ & FID $\downarrow$ & FID+ $\downarrow$ & Trans (cm) $\downarrow$ & AP $\uparrow$ & ROC-AUC $\uparrow$\\
        \midrule
        Full model & \textbf{0.874} & \textbf{0.769} & \textbf{0.027} & \textbf{0.603} & \textbf{0.653} & \textbf{0.374} & \textbf{0.553} & \textbf{0.017} & 0.192 & 1.31 & \textbf{0.832} & \textbf{0.772}\\
        w/o online multi-track attention & 0.851 & 0.764 & 0.048 & 0.552 & 0.642 & 0.372 & 0.551 & 0.046 & \textbf{0.179} & \textbf{1.14} & 0.811 & 0.747\\
        \bottomrule
    \end{tabular}
    }
\end{table}

\subsection{Ablation of Auxiliary Body-Part Loss}
\label{app:part_loss}

We set the weight of $\mathcal{L}_{\mathrm{part}}$ in Eq.~\eqref{eq:full_loss} to zero, removing auxiliary supervision of body-latent flow velocity while retaining the part-aware branches and prediction heads.
Table~\ref{tab:ablation_part_loss} shows small decreases on \dataset and HumanML3D.
These results suggest that direct body-part supervision provides a modest benefit to semantic alignment and temporal adherence beyond the part-aware architecture itself.

\begin{table}[!t]
    \centering
    \caption{Effect of the auxiliary body-part loss on HumanML3D and \dataset.}
    \label{tab:ablation_part_loss}
    \small
    \setlength{\tabcolsep}{1.8pt}
    \resizebox{\textwidth}{!}{
    \begin{tabular}{lcccccccccccc}
        \toprule
        \multirow{3}{*}{Variant} & \multicolumn{3}{c}{HumanML3D} & \multicolumn{9}{c}{\dataset}\\
        \cmidrule(lr){2-4}\cmidrule(lr){5-13}
        & \multirow{2}{*}{R@3$_{32}$ $\uparrow$} & \multirow{2}{*}{M2T $\uparrow$} & \multirow{2}{*}{FID $\downarrow$}
        & \multicolumn{2}{c}{Sequence alignment} & \multicolumn{2}{c}{Body-part alignment} & \multicolumn{3}{c}{Motion quality} & \multicolumn{2}{c}{Temporal adherence}\\
        \cmidrule(lr){5-6}\cmidrule(lr){7-8}\cmidrule(lr){9-11}\cmidrule(lr){12-13}
        &  &  &  & R@3$_{32}$ $\uparrow$ & M2T $\uparrow$ & R@3$_{32}$ $\uparrow$ & M2T $\uparrow$ & FID $\downarrow$ & FID+ $\downarrow$ & Trans (cm) $\downarrow$ & AP $\uparrow$ & ROC-AUC $\uparrow$\\
        \midrule
        Full model & \textbf{0.874} & \textbf{0.769} & \textbf{0.027} & \textbf{0.603} & \textbf{0.653} & \textbf{0.374} & \textbf{0.553} & \textbf{0.017} & \textbf{0.192} & 1.31 & \textbf{0.832} & \textbf{0.772}\\
        w/o $\mathcal{L}_{\mathrm{part}}$ & 0.865 & 0.764 & 0.028 & 0.596 & 0.651 & 0.371 & 0.552 & 0.019 & 0.195 & \textbf{1.26} & 0.826 & 0.762\\
        \bottomrule
    \end{tabular}
    }
\end{table}

\subsection{Ablation of Root-Velocity Loss}
\label{app:root_loss}

We remove $\mathcal{L}_{\mathrm{root}}$ from Eq.~\eqref{eq:full_loss}, disabling auxiliary smooth L1 supervision of frame-to-frame root displacement and heading changes.
Root features remain supervised by the main flow objective.
To assess smoothness beyond instruction boundaries, Table~\ref{tab:ablation_root_loss} additionally reports root and body jitter over complete sequences.
Root jitter measures the magnitude of third-order finite differences of the global root trajectory, while body jitter measures the corresponding root-relative quantity averaged over the other 21 joints.
Removing the root loss increases root jitter from 1.417 to 1.757 and body jitter from 0.614 to 0.680.
On \dataset, whole-motion FID increases from 0.017 to 0.021, sequence R@3 decreases from 0.603 to 0.595.
HumanML3D FID is similar between the two variants (0.026 vs. 0.027).
These results indicate that root-velocity supervision improves sequence-wide smoothness by reducing both root and body jitter.

\begin{table}[!t]
    \centering
    \caption{Effect of the root-motion loss on HumanML3D and \dataset.}
    \label{tab:ablation_root_loss}
    \small
    \setlength{\tabcolsep}{1.8pt}
    \resizebox{\textwidth}{!}{
    \begin{tabular}{lcccccccccccccc}
        \toprule
        \multirow{3}{*}{Variant} & \multicolumn{3}{c}{HumanML3D} & \multicolumn{11}{c}{\dataset}\\
        \cmidrule(lr){2-4}\cmidrule(lr){5-15}
        & \multirow{2}{*}{R@3$_{32}$ $\uparrow$} & \multirow{2}{*}{M2T $\uparrow$} & \multirow{2}{*}{FID $\downarrow$}
        & \multicolumn{2}{c}{Sequence alignment} & \multicolumn{2}{c}{Body-part alignment} & \multicolumn{5}{c}{Motion quality} & \multicolumn{2}{c}{Temporal adherence}\\
        \cmidrule(lr){5-6}\cmidrule(lr){7-8}\cmidrule(lr){9-13}\cmidrule(lr){14-15}
        &  &  &  & R@3$_{32}$ $\uparrow$ & M2T $\uparrow$ & R@3$_{32}$ $\uparrow$ & M2T $\uparrow$ & FID $\downarrow$ & FID+ $\downarrow$ & Trans (cm) $\downarrow$ & Root jitter $\downarrow$ & Body jitter $\downarrow$ & AP $\uparrow$ & ROC-AUC $\uparrow$\\
        \midrule
        Full model & \textbf{0.874} & 0.769 & 0.027 & \textbf{0.603} & \textbf{0.653} & \textbf{0.374} & 0.553 & \textbf{0.017} & 0.192 & 1.31 & \textbf{1.417} & \textbf{0.614} & \textbf{0.832} & \textbf{0.772}\\
        w/o $\mathcal{L}_{\mathrm{root}}$ & 0.872 & \textbf{0.771} & \textbf{0.026} & 0.595 & \textbf{0.653} & 0.372 & \textbf{0.555} & 0.021 & \textbf{0.187} & \textbf{1.27} & 1.757 & 0.680 & 0.827 & 0.763\\
        \bottomrule
    \end{tabular}
    }
\end{table}

\section{Extension to Spatial Control}
\label{app:control_details}

We extend the text-conditioned generator to accept spatial targets alongside text instructions.
At latent position \(n\), we zero out unavailable target values and concatenate the masked, normalized targets with their binary availability masks to form \(\boldsymbol{\xi}^{\mathrm{ctrl}}_n\).
A zero-initialized, bias-free linear projection~\citep{zhang2023controlnet} adds \(W_{\mathrm{c}}\boldsymbol{\xi}^{\mathrm{ctrl}}_n\) to the initial global state before part-global fusion.
The adapted model retains the streaming sampler and supports text-only generation with empty spatial masks.

\subsection{Additional Qualitative Results}
\label{app:additional_qualitative}

\begin{figure}[t]
    \centering
    \includegraphics[width=\linewidth]{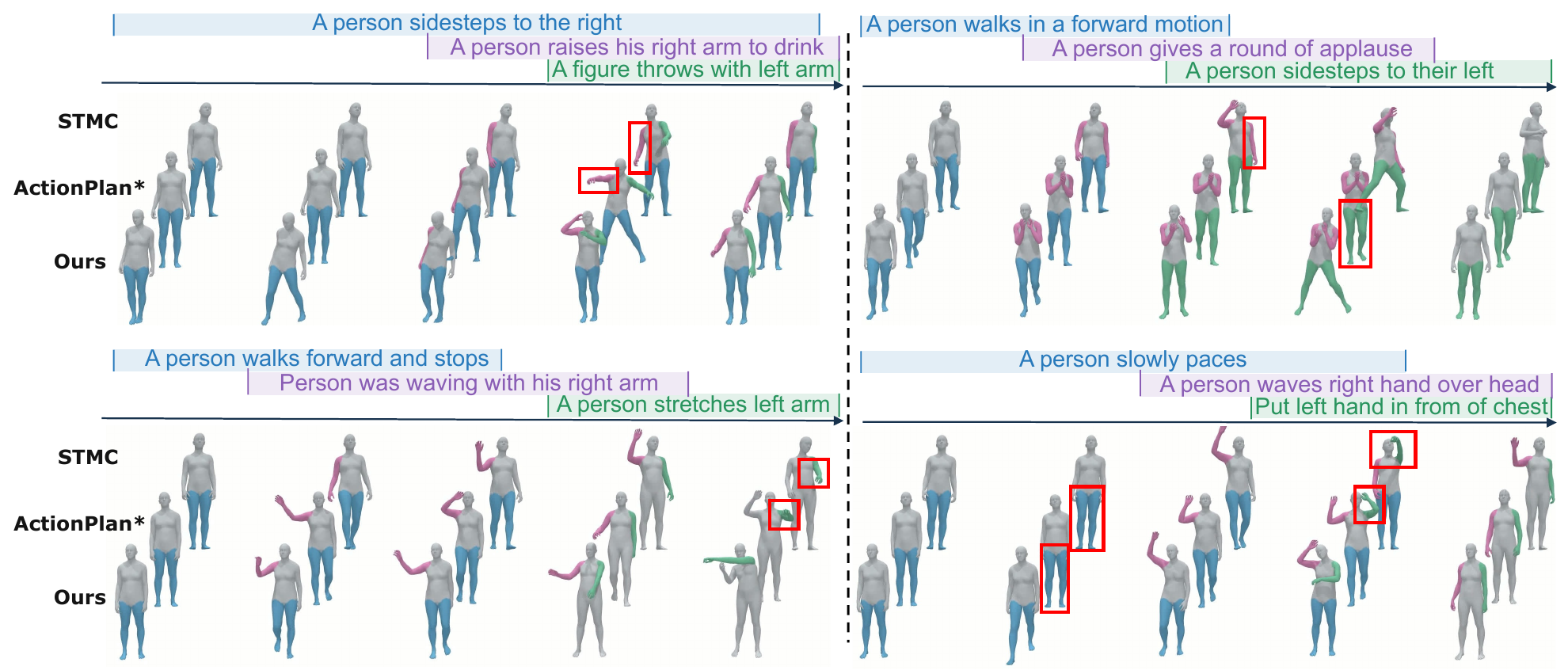}
    \caption{\textbf{Additional qualitative evaluation.} Red box marks the failure part.}
    
    \label{fig:additional_qualitative}
\end{figure}

Figure~\ref{fig:additional_qualitative} presents additional qualitative comparisons between \method, STMC, and ActionPlan* on multi-track motion generation. These examples complement Figure~\ref{fig:qualitative} in the main paper and illustrate model behavior under sequential and overlapping instructions.

\subsection{Spatial Conditions}

Spatial targets are represented in the root and body feature space with binary availability masks.
Only available controls on future frames are supplied, using five control types:
\begin{itemize}
    \item \emph{Root waypoints}: horizontal positions at sparse frames, optionally with heading.
    \item \emph{Root paths}: horizontal positions and heading at future frames with supplied root controls.
    \item \emph{Body keyframes}: root features and non-root joint positions at sparse frames.
    \item \emph{End-effector keyframes}: root features and selected foot or wrist positions and orientations.
    \item \emph{Foot contacts}: the four binary contact channels at selected frames.
\end{itemize}
Training samples position-only, orientation-only, or combined end-effector conditions. Evaluation uses combined conditions.
Combined conditioning uses the union of the masks for two distinct control types.

\subsection{Adaptation}

Spatial adaptation initializes from the exponential-moving-average weights of the trained generator.
We jointly fine-tune the denoiser and condition projection while keeping the autoencoder frozen, with learning rates \(2\times10^{-5}\) for the denoiser and \(10^{-4}\) for the condition projection.
Following the spatial-conditioning setup of ARDY~\citep{zhao2026ardy}, the objective is
\begin{equation}
    \mathcal{L}_{\mathrm{ctrl}}=\mathcal{L}_{\mathrm{gen}}+
      0.25\mathcal{L}_{\mathrm{dec}}
      +\mathcal{L}_{\mathrm{goal}}
      +\mathcal{L}_{\mathrm{kin}}.
    \label{eq:control_loss}
\end{equation}
The reconstruction term uses L1 error on decoded, normalized body features, while the goal term matches supplied root and body targets.
Kinematic consistency compares the explicit and forward-kinematic positions of the 21 non-root joints using mean Euclidean distance.
Training mixes single, combined, and absent spatial conditions, gradually increasing the maximum number of sparse keyframes.
Reconstruction and kinematic losses use valid future frames, while the goal loss applies only to supplied targets.

\subsection{Evaluation}

Each motion is randomly assigned to one of six control settings including five individual control types and one combined setting.
On \dataset, control availability is synchronized with text-prompt arrivals.
When a control becomes available, its world-space targets are translated by the horizontal displacement between the generated and source root positions at that time. The heading is unchanged.
Errors are measured relative to the transformed targets.
We report position errors for root, body, and end-effector targets in centimeters, and F1 for specified foot-contact labels.
Control metrics average applicable examples and evaluate only frames and joints with supplied targets.

\subsection{Results}

\begin{figure}[t]
    \centering
    \includegraphics[width=\textwidth]{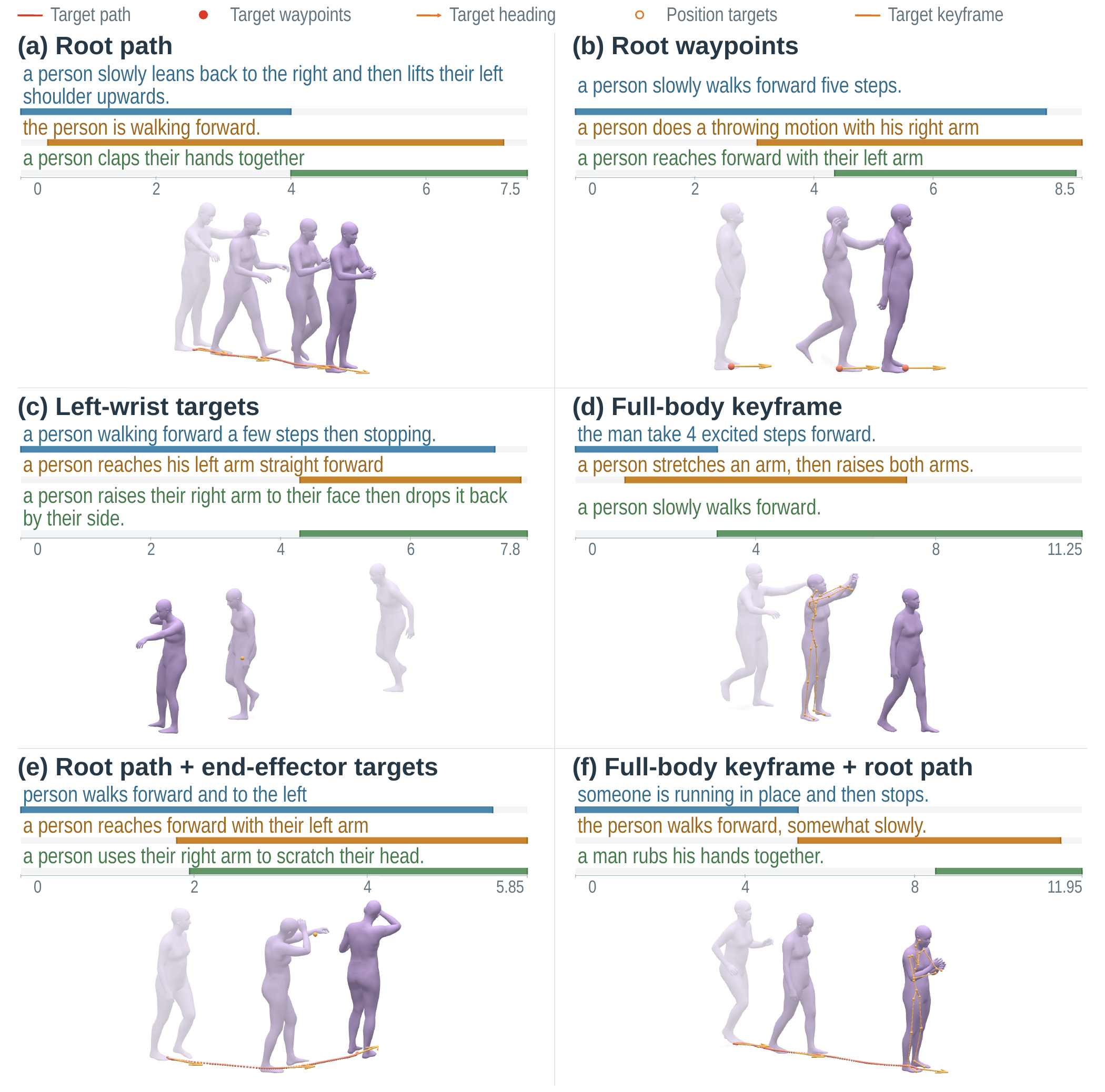}
    \caption{Motion generation with text and spatial controls: (a) root trajectories, (b) root waypoints, (c) left-wrist targets, (d) full-body keyframes, and root trajectories combined with (e) left-wrist targets or (f) full-body keyframes. Poses share a world coordinate frame within each panel and darken over time. Bars mark instruction intervals.}
    \label{fig:spatial_conditioning_additional}
\end{figure}

Table~\ref{tab:spatial_control} compares our extension with ARDY~\citep{zhao2026ardy}. Our extension achieves lower body and end-effector errors and higher contact F1 and M2T on both datasets, while ARDY remains more accurate for horizontal root targets.
Figure~\ref{fig:spatial_conditioning_additional} presents further examples with dense root trajectories, end-effector targets, and full-body keyframes.

\begin{table}[t]
    \centering
    \caption{Comparison with ARDY for motion generation conditioned jointly on text and spatial targets on HumanML3D and \dataset.}
    \label{tab:spatial_control}
    \small
    \setlength{\tabcolsep}{2.5pt}
    \resizebox{\textwidth}{!}{%
    \begin{tabular}{llccccccccc}
        \toprule
        \multirow{2}{*}{Testing Dataset} & \multirow{2}{*}{Method}
        & \multirow{2}{*}{Horizontal root (cm) $\downarrow$} & \multirow{2}{*}{Body (cm) $\downarrow$}
        & \multirow{2}{*}{End-effector (cm) $\downarrow$} & \multirow{2}{*}{Contact F1 $\uparrow$}
        & \multirow{2}{*}{M2T $\uparrow$} & \multirow{2}{*}{FID $\downarrow$} & \multirow{2}{*}{FID+ $\downarrow$}
        & \multicolumn{2}{c}{Temporal adherence}\\
        \cmidrule(lr){10-11}
        & & & & & & & & & AP $\uparrow$ & ROC-AUC $\uparrow$\\
        \midrule
        \multirow{2}{*}{HumanML3D} & ARDY & \textbf{2.55} & 8.15 & 14.06 & 0.857 & 0.650 & 0.246 & -- & -- & --\\
         & \method & 4.01 & \textbf{5.42} & \textbf{7.95} & \textbf{0.939} & \textbf{0.786} & \textbf{0.020} & -- & -- & --\\
        \midrule
        \multirow{2}{*}{\dataset} & ARDY & \textbf{4.13} & 9.98 & 16.88 & 0.897 & 0.565 & 0.221 & 0.308 & 0.715 & 0.630\\
         & \method & 6.31 & \textbf{8.05} & \textbf{12.63} & \textbf{0.942} & \textbf{0.644} & \textbf{0.013} & \textbf{0.216} & \textbf{0.820} & \textbf{0.759}\\
        \bottomrule
    \end{tabular}%
    }
\end{table}

\section{Robot Motion Execution}
\label{app:robot_execution}

\begin{figure}[t]
    \centering
    \includegraphics[width=\textwidth]{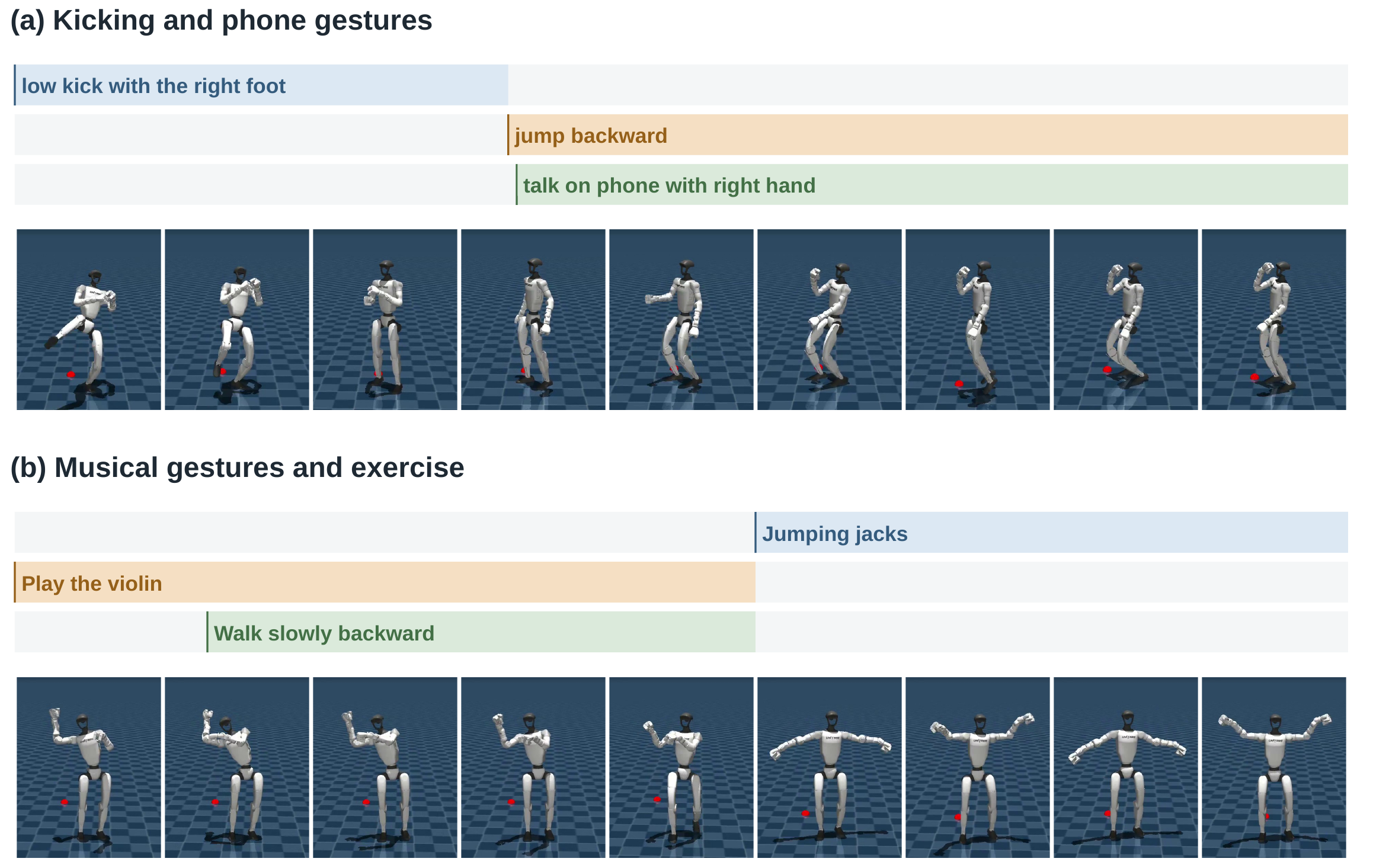}
    \caption{Text-driven G1 simulation. (a) A right-foot kick followed by overlapping instructions to jump backward and talk on the phone with the right hand. (b) Overlapping violin-playing and backward-walking instructions followed by jumping jacks. Bars indicate instruction intervals.}
    \label{fig:g1_motion_examples}
\end{figure}

\begin{figure}[t]
    \centering
    \includegraphics[width=\textwidth]{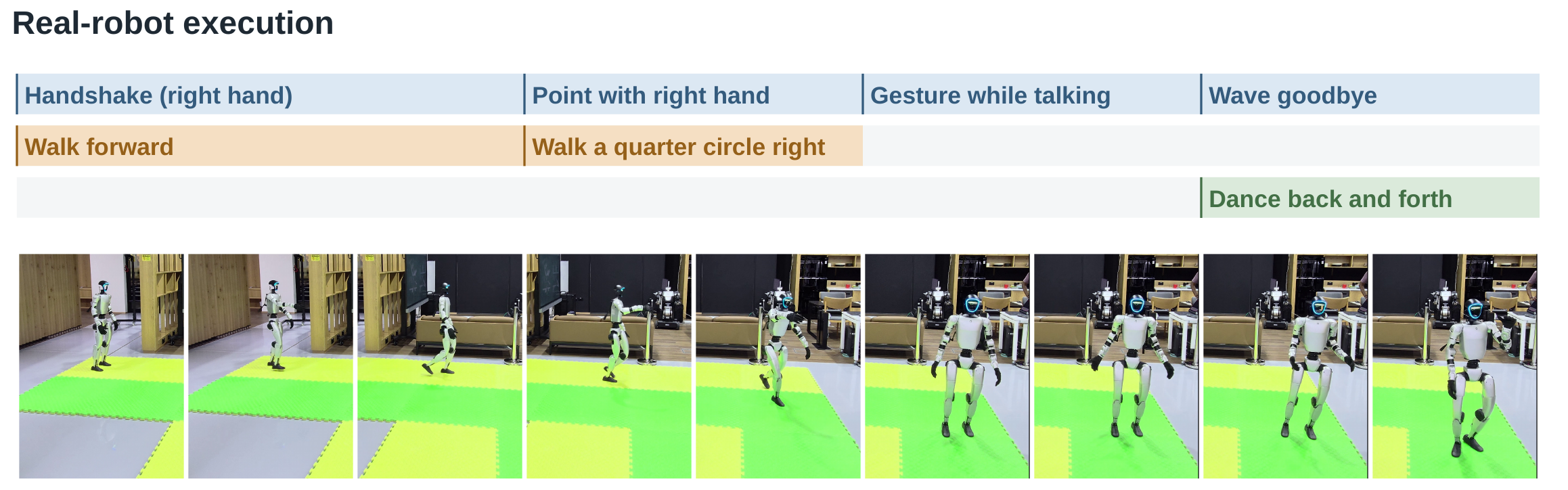}
    \caption{Text-driven real-robot execution combining upper-body gestures with walking and dancing. Bars indicate instruction intervals.}
    \label{fig:robot_execution_additional}
\end{figure}

We execute the generated motions on a physical humanoid robot in real time using SONIC's motion-tracking controller~\citep{luo2026sonic}.
Figure~\ref{fig:teaser}(c) shows upper-body gestures accompanying walking before a transition to a bow on the robot. The additional sequence in Figure~\ref{fig:robot_execution_additional} combines hand gestures with locomotion and stepping.
Figure~\ref{fig:g1_motion_examples} presents text-driven G1 motions in simulation, including kicking, phone gestures, and violin playing.

\section{Limitations}
\label{app:limitations}

\method has several limitations.
First, \dataset is constructed by composing HumanML3D source motions rather
than collecting motions from real streaming multi-track interactions. Although we
filter incompatible actions and refine composition boundaries, the resulting data
may not capture the full diversity and complexity of naturally co-occurring actions.
Second, our formulation uses explicit temporal intervals to determine when each
instruction is active. When an end time is not provided, the system can use a
fixed-duration interval (e.g., 5 seconds) and extend the instruction in successive
intervals for continuous execution. However, it does not currently infer action
duration or termination directly from language or motion state.
Finally, our part-aware supervision relies on automatically generated body-part
annotations and a fixed decomposition into six anatomical regions; annotation errors
or actions involving more complex cross-part interactions may therefore limit the
part-aware guidance. Extending the framework to naturally collected interactive
data, learned action termination, and more flexible part grounding provides
promising directions for future work.

\end{document}